\documentclass[final,10pt,3p]{elsarticle}

\usepackage[utf8]{inputenc}
\usepackage{amssymb}
\usepackage{bm}
\usepackage{algorithm}
\usepackage{algpseudocode}
\usepackage{blindtext}

\usepackage{graphicx}
\usepackage{subcaption}
\usepackage{multirow}

\usepackage[dvipsnames]{xcolor}
\usepackage{amsmath}
\usepackage{tikz}
\usepackage{mathdots}
\usepackage{yhmath}
\usepackage{cancel}
\usepackage{color}
\usepackage{siunitx}
\usepackage{array}
\usepackage{multirow}
\usepackage{amssymb}
\usepackage{gensymb}
\usepackage{tabularx}
\usepackage{booktabs}
\usetikzlibrary{fadings}
\usepackage{gensymb}
\usepackage{hyperref}

\usepackage{mathtools}
\DeclarePairedDelimiter\norm{\lVert}{\rVert}%

\journal{Information Fusion}

\begin{document}

\begin{frontmatter}



\title{Textual Data for Time Series Forecasting}

\author[label1,label2]{David Obst}
\ead{david.obst@edf.fr}
\author[label2]{Badih Ghattas}
\ead{badih.ghattas@univ-amu.fr}
\author[label1]{Sandra Claudel}
\ead{sandra.claudel@edf.fr}
\author[label3]{Jairo Cugliari}
\ead{Jairo.Cugliari@univ-lyon2.fr}
\author[label1]{Yannig Goude}
\ead{yannig.goude@edf.fr}
\author[label4]{Georges Oppenheim}
\ead{georges.oppenheim@etud.u-pem.fr}

\address[label1]{EDF R\&D, Palaiseau, France } 
\address[label2]{Institut de Mathématiques de Marseille, Aix-Marseille Université, France } 
\address[label3]{ERIC, Université de Lyon 2, France} 
\address[label4]{Laboratoire d’Analyse et de Mathématiques Appliquées Université Paris-Est, Champs-sur-Marne, France}




\begin{abstract}

\noindent While ubiquitous, textual sources of information such as company reports, social media posts, etc. are hardly included in prediction algorithms for time series, despite the relevant information they may contain. In this work, openly accessible daily weather reports from France and the United-Kingdom are leveraged to predict time series of national electricity consumption, average temperature and wind-speed with a single pipeline. Two methods of numerical representation of text are considered, namely traditional Term Frequency - Inverse Document Frequency (TF-IDF) as well as our own neural word embedding. Using exclusively text, we are able to predict the aforementioned time series with sufficient accuracy to be used to replace missing data. Furthermore the proposed word embeddings display geometric properties relating to the behavior of the time series and context similarity between words.

\end{abstract}


\end{frontmatter}


\section{Introduction}
\label{Sec:Introduction}

Whether it is in the field of energy, finance or meteorology, accurately predicting the behavior of time series is nowadays of paramount importance for optimal decision making or profit. While the field of time series forecasting is extremely prolific from a research point-of-view, up to now it has narrowed its efforts on the exploitation of regular numerical features extracted from sensors, data bases or stock exchanges. Unstructured data such as text on the other hand remains underexploited for prediction tasks, despite its potentially valuable informative content. Empirical studies have already proven that textual sources such as news articles or blog entries can be correlated to stock exchange time series and have explanatory power for their variations \cite{ruiz2012correlating,chen2014wisdom}. This observation has motivated multiple extensive experiments to extract relevant features from textual documents in different ways and use them for prediction, notably in the field of finance. In Lavrenko et al. \cite{lavrenko2000mining}, language models (considering only the presence of a word) are used to estimate the probability of trends such as surges or falls of 127 different stock values using articles from Biz Yahoo!. Their results show that this text driven approach could be used to make profit on the market. One of the most conventional ways for text representation is the TF-IDF (Term Frequency - Inverse Document Frequency) approach. Authors have included such features derived from news pieces in multiple traditional machine learning algorithms such as support vector machines (SVM) \cite{fung2005predicting} or logistic regression \cite{wang2017study} to predict the variations of financial series again. An alternative way to encode the text is through latent Dirichlet allocation (LDA) \cite{blei2003latent}. It assigns topic probabilities to a text, which can be used as inputs for subsequent tasks. This is for instance the case in Wang's aforementioned work (alongside TF-IDF). In \cite{atkins2018financial}, the authors used Reuters news encoded by LDA to predict if NASDAQ and Dow Jones closing prices increased or decreased compared to the opening ones. Their empirical results show that this approach was efficient to improve the prediction of stock volatility. More recently Kanungsukkasem et al. \cite{kanungsukkasem2019financial} introduced a variant of the LDA graphical model, named FinLDA, to craft probabilities that are specifically tailored for a financial time series prediction task (although their approach could be generalized to other ones). Their results showed that indeed performance was better when using probabilities from their alternative than those of the original LDA. Deep learning with its natural ability to work with text through word embeddings has also been used for time series prediction with text. Combined with traditional time series features, the authors of \cite{li2018text} derived sentiment features from a convolutional neural network (CNN) to reduce the prediction error of oil prices. Akita et al. \cite{akita2016deep} represented news articles through the use of paragraph vectors \cite{dai2015document} in order to predict 10 closing stock values from the Nikkei 225. While in the case of financial time series the existence of specialized press makes it easy to decide \emph{which} textual source to use, it is much more tedious in other fields. Recently in Rodrigues et al. \cite{rodrigues2019combining}, short  description of events (such as concerts, sports matches, ...) are leveraged through a word embedding and neural networks in addition to more traditional features. Their experiments show that including the text can bring an improvement of up to 2\% of root mean squared error compared to an approach without textual information. Although the presented studies conclude on the usefulness of text to improve predictions, they never thoroughly analyze which aspects of the text are of importance, keeping the models as black-boxes.

The field of electricity consumption is one where expert knowledge is broad. It is known that the major phenomena driving the load demand are calendar (time of the year, day of the week, ...) and meteorological. For instance generalized additive models (GAM) \cite{wood2017generalized} representing the consumption as a sum of functions of the time of the year, temperature and wind speed (among others) typically yield less than 1.5\% of relative error for French national electricity demand and 8\% for local one \cite{pierrot2011short,goude2013local}. Neural networks and their variants, with their ability to extract patterns from heterogeneous types of data have also obtained state-of-the-art results \cite{park1991electric,li2015short,ryu2016deep}. However to our knowledge no exploratory work using text has been conducted yet. Including such data in electricity demand forecasting models would not only contribute to close the gap with other domains, but also help to understand better which aspects of text are useful, how the encoding of the text influences forecasts and to which extend a prediction algorithm can extract relevant information from unstructured data. Moreover the major drawback of all the aforementioned approaches is that they require meteorological data that may be difficult to find, unavailable in real time or expensive. Textual sources such as weather reports on the other hand are easy to find, usually available on a daily basis and free.

The main contribution of our paper is to suggest the use of a certain type of textual documents, namely daily weather report, to build forecasters of the daily national electricity load, average temperature and wind speed for both France and the United-Kingdom (UK). Consequently this work represents a significant break with traditional methods, and we do not intend to best state-of-the-art approaches. Textual information is naturally more fuzzy than numerical one, and as such the same accuracy is not expected from the presented approaches. With a single text, we were already able to predict the electricity consumption with a relative error of less than 5\% for both data sets. Furthermore, the quality of our predictions of temperature and wind speed is satisfying enough to replace missing or unavailable data in traditional models. Two different approaches are considered to represent the text numerically, as well as multiple forecasting algorithms. Our empirical results are consistent across encoding, methods and language, thus proving the intrinsic value weather reports have for the prediction of the aforementioned time series. Moreover, a major distinction between previous works is our interpretation of the models. We quantify the impact of a word on the forecast and analyze the geometric properties of the word embedding we trained ourselves. Note that although multiple time series are discussed in our paper, the main focus of this paper remains electricity consumption. As such, emphasis is put on the predictive results on the load demand time series.

The rest of this paper is organized as follows. The following section introduces the two data sets used to conduct our study. Section 3 presents the different machine learning approaches used and how they were tuned. Section 4 highlights the main results of our study, while section 5 concludes this paper and gives insight on future possible work.

\section{Presentation of the data}
\label{Sec:PresentationData}

In order to prove the consistency of our work, experiments have been conducted on two data sets, one for France and the other for the UK. In this section details about the text and time series data are given, as well as the major preprocessing steps.

\subsection{Time Series}

\begin{figure}[H]
    \centering
    \begin{subfigure}[b]{0.49\textwidth}
        \includegraphics[width=\textwidth]{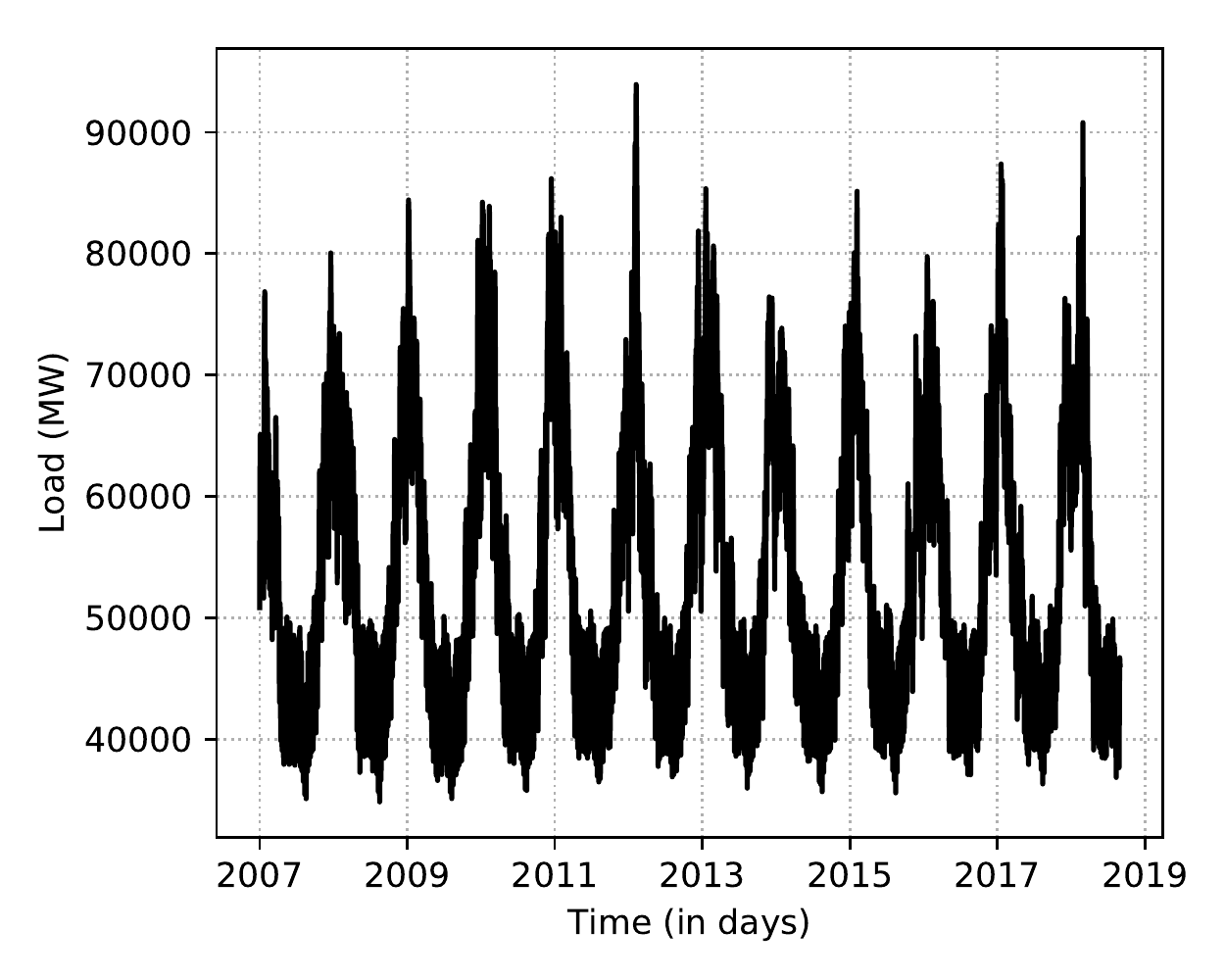}
        \caption{France}
        \label{Fig:load_FR_trend}
    \end{subfigure}
    \hspace{-0.2cm} 
    \begin{subfigure}[b]{0.49\textwidth}
        \includegraphics[width=\textwidth]{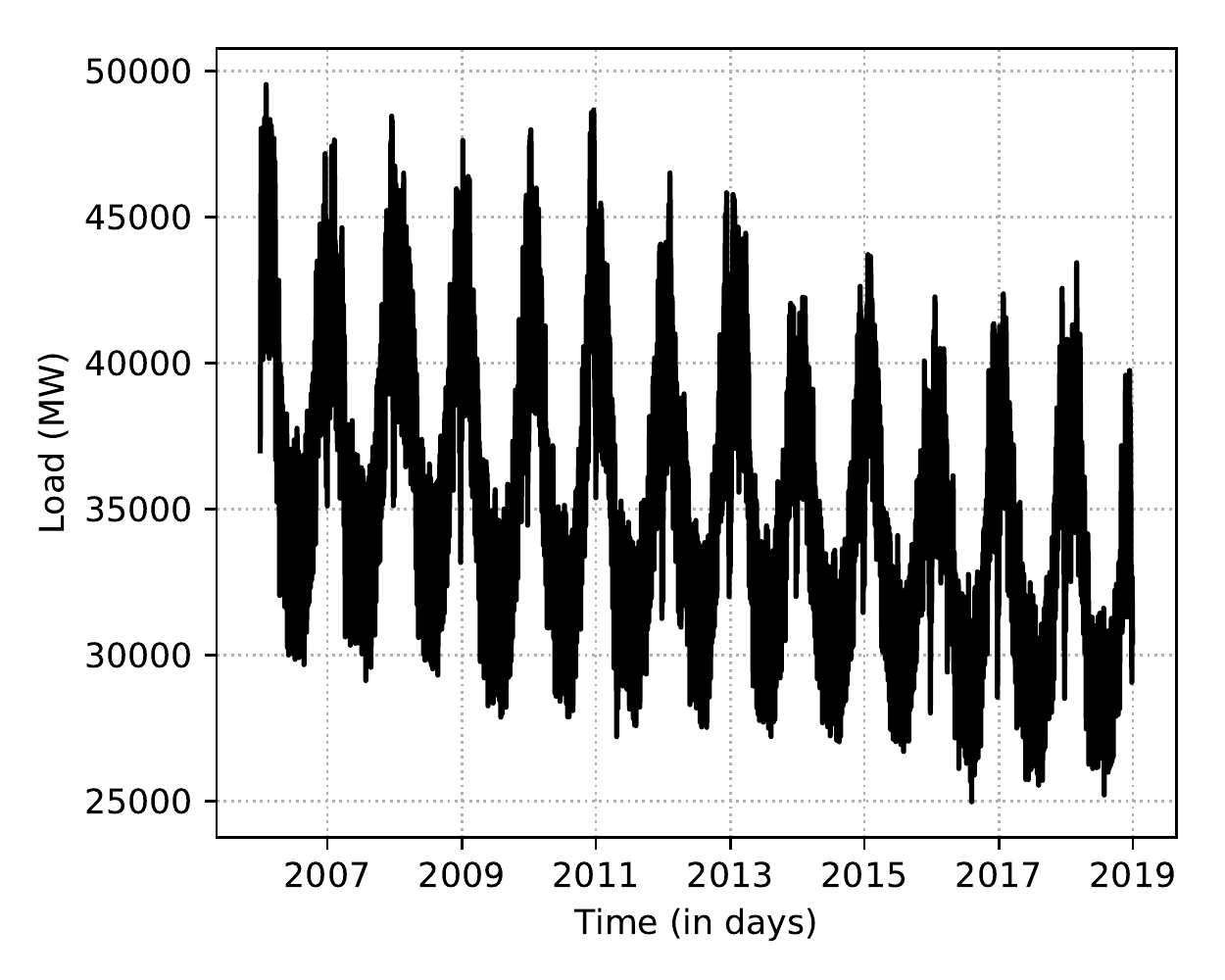}
        \caption{UK}
        \label{Fig:load_UK_trend}
    \end{subfigure}
    \caption{Net electricity consumption (Load) over time.} \label{Fig:loads}
\end{figure}

Three types of time series are considered in our work: national net electricity consumption (also referred as load or demand), national temperature and wind speed. The load data sets were retrieved on the websites of the respective grid operators, respectively RTE (Réseau et Transport d'\'{E}lectricité) for France and National Grid for the UK. For France, the available data ranges from January the 1\textsuperscript{st} 2007 to August the 31\textsuperscript{st} 2018. The default temporal resolution is 30 minutes, but it is averaged to a daily one. For the UK, it is available from January the 1\textsuperscript{st} 2006 to December the 31\textsuperscript{st} 2018 with the same temporal resolution and thus averaging. Due to social factors such as energy policies or new usages of electricity (e.g. Electric Vehicles), the net consumption usually has a long-term trend (fig. \ref{Fig:loads}). While for France it seems marginal (fig. \ref{Fig:load_FR_trend}), there is a strong decreasing trend for the United-Kingdom (fig. \ref{Fig:load_UK_trend}). Such a strong non-stationarity of the time series would cause problems for the forecasting process, since the learnt demand levels would differ significantly from the upcoming ones. Therefore a linear regression was used to approximate the decreasing trend of the net consumption in the UK. It is then subtracted before the training of the methods, and then re-added a posteriori for prediction. 


As for the weather time series, they were extracted from multiple weather stations around France and the UK. The national average is obtained by combining the data from all stations with a weight proportional to the city population the station is located in. For France the stations' data is provided by the French meteorological office, Météo France, while the British ones are scrapped from stations of the National Oceanic and Atmospheric Administration (NOAA). Available on the same time span as the consumption, they usually have a 3 hours temporal resolution but are averaged to a daily one as well. Finally the time series were scaled to the range $[0,1]$ before the training phase, and re-scaled during prediction time.

\subsection{Text}

Our work aims at predicting time series using exclusively text. Therefore for both countries the inputs of all our models consist only of written daily weather reports. Under their raw shape, those reports take the form of PDF documents giving a short summary of the country's overall weather, accompanied by pressure, temperature, wind, etc. maps. Note that those reports are written \textit{a posteriori}, although they could be written in a predictive fashion as well. The reports are published by Météo France and the Met Office, its British counterpart. They are publicly available on the respective websites of the organizations. Both corpora span on the same period as the corresponding time series and given their daily nature, it yields a total of 4,261 and 4,748 documents respectively. An excerpt for each language may be found in tables \ref{Tab:excerpt_report_UK} and \ref{Tab:excerpt_report_FR}. The relevant text was extracted from the PDF documents using the Python library \texttt{PyPDF2}. 

\begin{figure}[H]
    \centering
    \begin{subfigure}[b]{0.47\textwidth}
        \includegraphics[width=\textwidth]{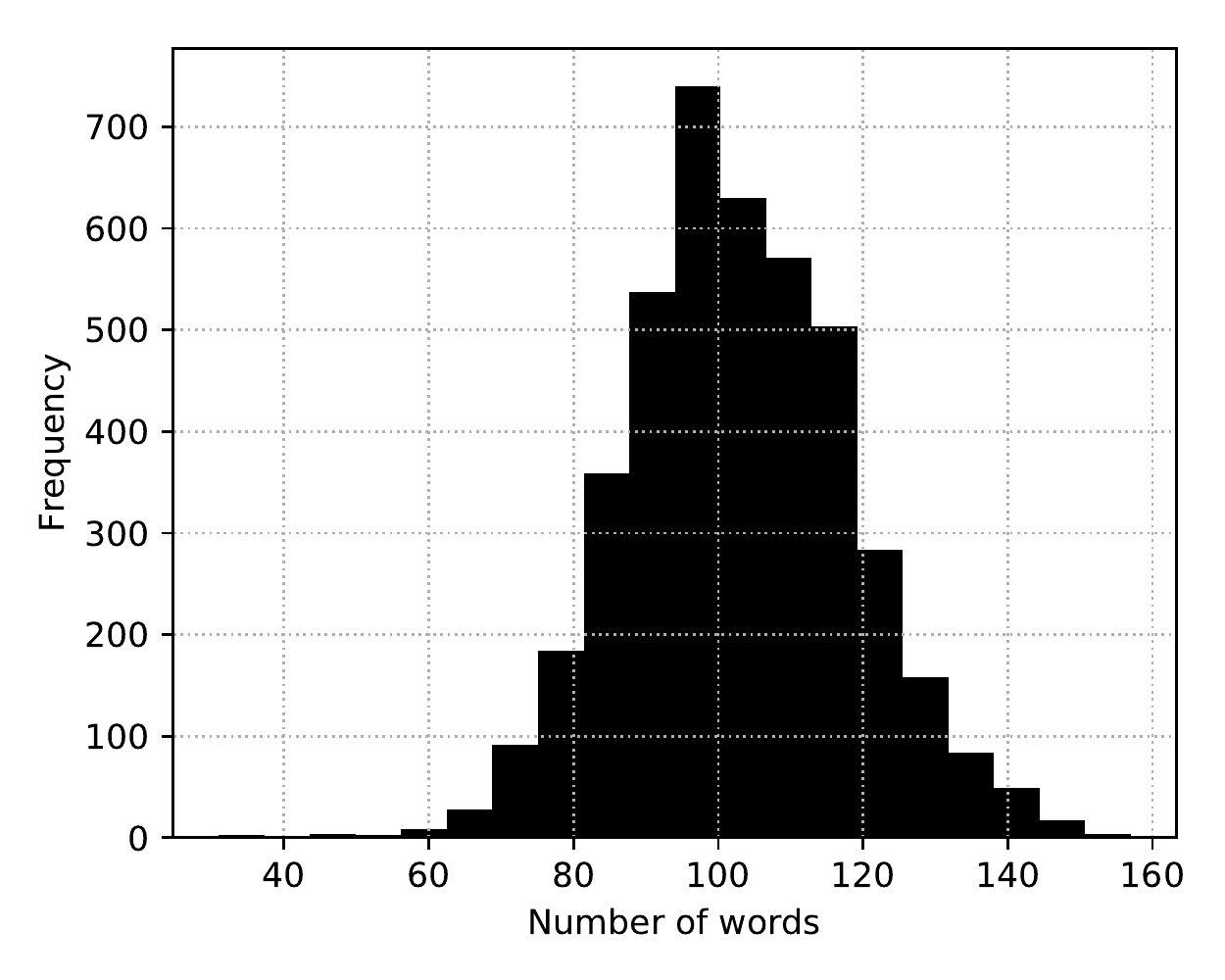}
        \caption{French reports}
    \end{subfigure}
    \hspace{-0.2cm} 
    \begin{subfigure}[b]{0.47\textwidth}
        \includegraphics[width=\textwidth]{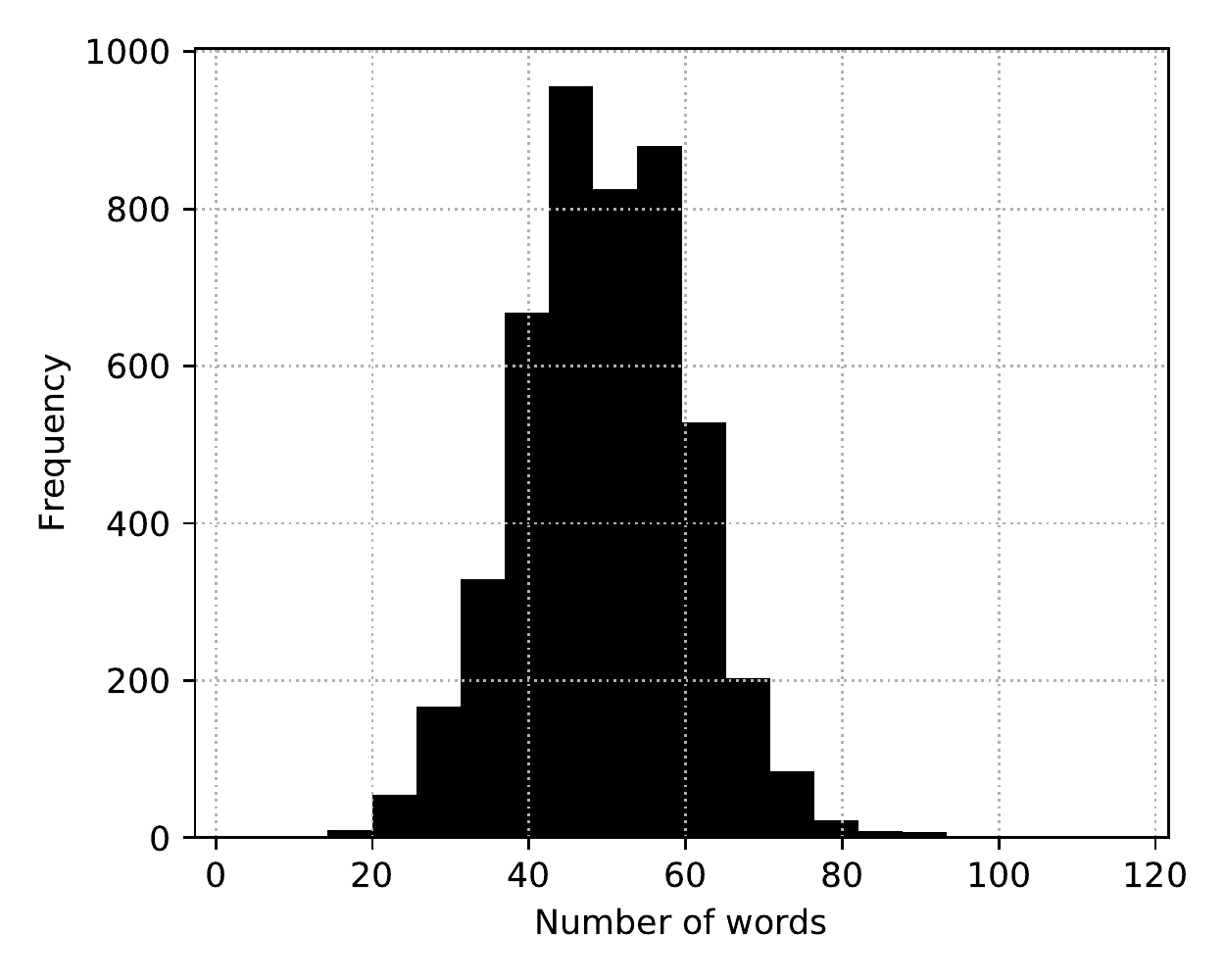}
        \caption{UK reports}
    \end{subfigure}
    \caption{Word counts for the two corpora after preprocessing.} \label{Fig:histo_words}
\end{figure}

As emphasized in many studies, preprocessing of the text can ease the learning of the methods and improve accuracy \cite{uysal2014impact}. Therefore the following steps are applied: removal of non-alphabetic characters, removal of stop-words and lowercasing. While it was often highlighted that word lemmatization and stemming improve results, initial experiments showed it was not the case for our study. This is probably due to the technical vocabulary used in both corpora pertaining to the field of meteorology. Already limited in size, the aforementioned preprocessing operations do not yield a significant vocabulary size reduction and can even lead to a loss of linguistic meaning. Finally, extremely frequent or rare words may not have high explanatory power and may reduce the different models' accuracy. That is why words appearing less than 7 times or in more than 40\% of the (learning) corpus are removed as well. Figure \ref{Fig:histo_words} represents the distribution of the document lengths after preprocessing, while table \ref{Tab:corpora_descript} gives descriptive statistics on both corpora. Note that the preprocessing steps do not heavily rely on the considered language: therefore our pipeline is easily adaptable for other languages.

\begin{table}[H]
    \centering
    \begin{tabular}{l}
    \hline
        Summary of UK Weather for Monday 01 February 2016. \\
        \\
        Overnight, rain pushed northwards across Scotland. This introduced milder conditions across \\ the whole of the U.K.,so it was a mild, cloudy night with drizzle and hill fog in the west [...]\\
    \hline
    \end{tabular}
    \caption{Excerpt of the Met Office report of Monday the 1\textsuperscript{st} February 2016.}
    \label{Tab:excerpt_report_UK}
\end{table}

\begin{table}[H]
    \centering
    \begin{tabular}{l}
    \hline
        Mercredi 4 janvier 2017 \\
        \\
        En altitude, un thalweg bien froid et dynamique se décale sur 
        le flanc est de la France.[...] Le matin, des \\ 
        neiges faibles glissent le long des frontières du Nord et atteignent le Nord-Est. De la pluie verglaçante [...] \\
    \hline
    \end{tabular}
    \caption{Excerpt of the Météo France report of Wednesday the 4\textsuperscript{th} January 2017.}
    \label{Tab:excerpt_report_FR}
\end{table}

\begin{table}[h]
    \centering
    \begin{tabular}{|c|c|c|c|}
    \hline
        Corpus & Vocabulary Size & Max sentence length & Average sentence length \\
        \hline
        France & 3312 & 157 & 103 \\
        \hline
        UK & 1202 & 116 & 50 \\
    \hline
    \end{tabular}
    \caption{Descriptive analysis of the two corpora (after preprocessing)}
    \label{Tab:corpora_descript}
\end{table}

\section{Modeling and forecasting framework}
\label{Sec:ModelingAndForecastingFramework}

A major target of our work is to show the reports contain an intrinsic information relevant for time series, and that the predictive results do not heavily depend on the encoding of the text or the machine learning algorithm used. Therefore in this section we present the text encoding approaches, as well as the forecasting methods used with them.

\subsection{Numerical Encoding of the Text}

Machines and algorithms cannot work with raw text directly. Thus one major step when working with text is the choice of its numerical representation. In our work two significantly different encoding approaches are considered. The first one is the TF-IDF approach. It embeds a corpus of $N$ documents and $V$ words into a matrix $X$ of size $N \times V$. As such, every document is represented by a vector of size $V$. For each word $w$ and document $d$ the associated coefficient $x_{d,w}$ represents the frequency of that word in that document, penalized by its overall frequency in the rest of the corpus. Thus very common words will have a low TF-IDF value, whereas specific ones which will appear often in a handful of documents will have a large TF-IDF score. The exact formula to calculate the TF-IDF value of word $w$ in document $d$ is:

$$
x_{d,w} = f_{d,w} \times \ln \left( \frac{N}{\#\{d: w \in d \} + 1} \right)
$$

\noindent where $f_{d,w}$ is the number of appearances of $w$ in $d$ adjusted by the length of $d$ and $\#\{d: w \in d \}$ is the number of documents in which the word $w$ appears. In our work we considered only individual words, also commonly referred as 1-grams in the field of natural language processing (NLP). The methodology can be easily extended to $n$-grams (groups of $n$ consecutive words), but initial experiments showed that it did not bring any significant improvement over 1-grams.

The second representation is a neural word embedding. It consists in representing every word in the corpus by a real-valued vector of dimension $q$. Such models are usually obtained by learning a vector representation from word co-occurrences in a very large corpus (typically hundred thousands of documents, such as Wikipedia articles for example). The two most popular embeddings are probably Google's Word2Vec \cite{word2vec} and Standford's GloVe \cite{GloVe}. In the former, a neural network is trained to predict a word given its context (continuous bag of word model), whereas in the latter a matrix factorization scheme on the log co-occurences of words is applied. In any case, the very nature of the objective function allows the embedding models to learn to translate linguistic similarities into geometric properties in the vector space. For instance the vector $\overrightarrow{king} - \overrightarrow{man} + \overrightarrow{woman}$ is expected to be very close to the vector $\overrightarrow{queen}$. However in our case we want a vector encoding which is tailored for the technical vocabulary of our weather reports and for the subsequent prediction task. This is why we decided to train our own word embedding from scratch during the learning phase of our recurrent or convolutional neural network. Aside from the much more restricted size of our corpora, the major difference with the aforementioned embeddings is that in our case it is obtained by minimizing a squared loss on the prediction. In that framework there is no explicit reason for our representation to display any geometric structure. However as detailed in section \ref{Sec:embedding}, our word vectors nonetheless display geometric properties pertaining to the behavior of the time series.

\subsection{Machine Learning Algorithms}

Multiple machine learning algorithms were applied on top of the encoded textual documents. For the TF-IDF representation, the following approaches are applied: random forests (RF), LASSO and multilayer perceptron (MLP) neural networks (NN). We chose these algorithms combined to the TF-IDF representation due to the possibility of interpretation they give. Indeed, considering the novelty of this work, the understanding of the impact of the words on the forecast is of paramount importance, and as opposed to embeddings, TF-IDF has a natural interpretation. Furthermore the RF and LASSO methods give the possibility to interpret marginal effects and analyze the importance of features, and thus to find the words which affect the time series the most.

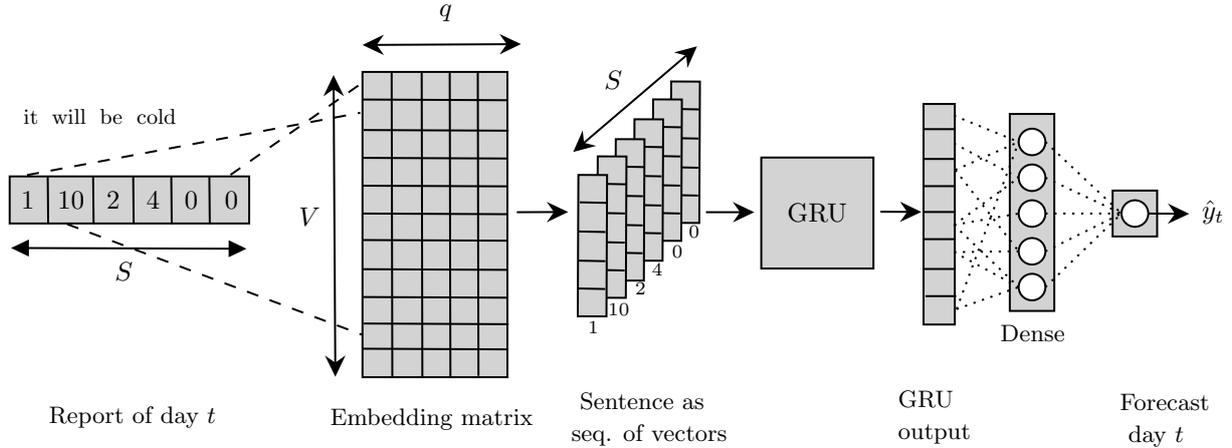
\begin{figure}[H]
    \centering
    
\tikzset{every picture/.style={line width=0.75pt}} 

\begin{tikzpicture}[x=0.75pt,y=0.75pt,yscale=-1,xscale=1]

\draw  [fill={rgb, 255:red, 211; green, 211; blue, 211 }  ,fill opacity=1 ] (569.89,115.12) -- (592.11,115.12) -- (592.11,138.21) -- (569.89,138.21) -- cycle ;
\draw  [fill={rgb, 255:red, 211; green, 211; blue, 211 }  ,fill opacity=1 ] (363.88,60.31) -- (363.57,131.42) -- (349.23,131.36) -- (349.54,60.25) -- cycle ;
\draw    (363.82,74.31) -- (350.26,74.03) ;

\draw    (363.09,88.97) -- (349.53,88.69) ;

\draw    (363.02,103.64) -- (349.47,103.36) ;

\draw    (362.96,117.64) -- (349.41,117.36) ;

\draw  [fill={rgb, 255:red, 211; green, 211; blue, 211 }  ,fill opacity=1 ] (354.88,69.31) -- (354.57,140.42) -- (340.23,140.36) -- (340.54,69.25) -- cycle ;
\draw    (354.82,83.31) -- (341.26,83.03) ;

\draw    (354.09,97.97) -- (340.53,97.69) ;

\draw    (354.02,112.64) -- (340.47,112.36) ;

\draw    (353.96,126.64) -- (340.41,126.36) ;

\draw  [fill={rgb, 255:red, 211; green, 211; blue, 211 }  ,fill opacity=1 ] (345.54,79.31) -- (345.23,150.42) -- (330.9,150.36) -- (331.21,79.25) -- cycle ;
\draw    (345.48,93.31) -- (331.93,93.03) ;

\draw    (344.75,107.97) -- (331.2,107.69) ;

\draw    (344.69,122.64) -- (331.13,122.36) ;

\draw    (344.63,136.64) -- (331.07,136.36) ;

\draw  [fill={rgb, 255:red, 211; green, 211; blue, 211 }  ,fill opacity=1 ] (336.21,89.31) -- (335.9,160.42) -- (321.57,160.36) -- (321.88,89.25) -- cycle ;
\draw    (336.15,103.31) -- (322.6,103.03) ;

\draw    (335.42,117.97) -- (321.86,117.69) ;

\draw    (335.35,132.64) -- (321.8,132.36) ;

\draw    (335.29,146.64) -- (321.74,146.36) ;

\draw  [fill={rgb, 255:red, 211; green, 211; blue, 211 }  ,fill opacity=1 ] (327.21,97.98) -- (326.9,169.09) -- (312.57,169.02) -- (312.88,97.91) -- cycle ;
\draw    (327.15,111.98) -- (313.6,111.69) ;

\draw    (326.42,126.64) -- (312.86,126.36) ;

\draw    (326.35,141.31) -- (312.8,141.02) ;

\draw    (326.29,155.31) -- (312.74,155.02) ;

\draw  [fill={rgb, 255:red, 211; green, 211; blue, 211 }  ,fill opacity=1 ] (195.44,55.44) -- (267.78,55.44) -- (267.78,209) -- (195.44,209) -- cycle ;
\draw    (195.33,194.67) -- (267.78,195) ;

\draw    (181.01,59) -- (181.88,206.78) ;
\draw [shift={(181.89,208.78)}, rotate = 269.65999999999997] [fill={rgb, 255:red, 0; green, 0; blue, 0 }  ][line width=0.75]  [draw opacity=0] (10.72,-5.15) -- (0,0) -- (10.72,5.15) -- (7.12,0) -- cycle    ;
\draw [shift={(181,57)}, rotate = 89.66] [fill={rgb, 255:red, 0; green, 0; blue, 0 }  ][line width=0.75]  [draw opacity=0] (10.72,-5.15) -- (0,0) -- (10.72,5.15) -- (7.12,0) -- cycle    ;
\draw    (197.44,41.88) -- (272.56,41.46) ;
\draw [shift={(274.56,41.44)}, rotate = 539.6800000000001] [fill={rgb, 255:red, 0; green, 0; blue, 0 }  ][line width=0.75]  [draw opacity=0] (10.72,-5.15) -- (0,0) -- (10.72,5.15) -- (7.12,0) -- cycle    ;
\draw [shift={(195.44,41.89)}, rotate = 359.68] [fill={rgb, 255:red, 0; green, 0; blue, 0 }  ][line width=0.75]  [draw opacity=0] (10.72,-5.15) -- (0,0) -- (10.72,5.15) -- (7.12,0) -- cycle    ;
\draw    (302.13,94.42) -- (360.93,43.86) ;
\draw [shift={(362.44,42.56)}, rotate = 499.31] [fill={rgb, 255:red, 0; green, 0; blue, 0 }  ][line width=0.75]  [draw opacity=0] (10.72,-5.15) -- (0,0) -- (10.72,5.15) -- (7.12,0) -- cycle    ;
\draw [shift={(300.61,95.72)}, rotate = 319.31] [fill={rgb, 255:red, 0; green, 0; blue, 0 }  ][line width=0.75]  [draw opacity=0] (10.72,-5.15) -- (0,0) -- (10.72,5.15) -- (7.12,0) -- cycle    ;
\draw  [fill={rgb, 255:red, 211; green, 211; blue, 211 }  ,fill opacity=1 ] (394.5,98.5) -- (450.5,98.5) -- (450.5,154.33) -- (394.5,154.33) -- cycle ;
\draw  [fill={rgb, 255:red, 211; green, 211; blue, 211 }  ,fill opacity=1 ] (475.5,71.56) -- (491.11,71.56) -- (491.11,182.5) -- (475.5,182.5) -- cycle ;
\draw  [fill={rgb, 255:red, 211; green, 211; blue, 211 }  ,fill opacity=1 ] (518.56,76.58) -- (540.78,76.58) -- (540.78,175.67) -- (518.56,175.67) -- cycle ;
\draw  [fill={rgb, 255:red, 211; green, 211; blue, 211 }  ,fill opacity=1 ] (19.33,108) -- (138.89,108) -- (138.89,131.67) -- (19.33,131.67) -- cycle ;
\draw    (272.5,126) -- (296.11,125.9) ;
\draw [shift={(298.11,125.89)}, rotate = 539.75] [fill={rgb, 255:red, 0; green, 0; blue, 0 }  ][line width=0.75]  [draw opacity=0] (10.72,-5.15) -- (0,0) -- (10.72,5.15) -- (7.12,0) -- cycle    ;

\draw    (366.89,126) -- (389,126) ;
\draw [shift={(391,126)}, rotate = 180] [fill={rgb, 255:red, 0; green, 0; blue, 0 }  ][line width=0.75]  [draw opacity=0] (10.72,-5.15) -- (0,0) -- (10.72,5.15) -- (7.12,0) -- cycle    ;

\draw    (454,125.75) -- (472.44,125.88) ;
\draw [shift={(474.44,125.89)}, rotate = 180.39] [fill={rgb, 255:red, 0; green, 0; blue, 0 }  ][line width=0.75]  [draw opacity=0] (10.72,-5.15) -- (0,0) -- (10.72,5.15) -- (7.12,0) -- cycle    ;

\draw  [dash pattern={on 4.5pt off 4.5pt}]  (28,107.33) -- (194.33,76) ;

\draw  [dash pattern={on 4.5pt off 4.5pt}]  (48,132) -- (195.33,187.67) ;

\draw    (38.33,108.33) -- (38.56,131.89) ;

\draw    (61,108.33) -- (61.22,131.89) ;

\draw    (81,108.33) -- (81.22,131.89) ;

\draw    (100,108.33) -- (100.22,131.89) ;

\draw    (119,108.33) -- (119.22,131.89) ;

\draw  [dash pattern={on 4.5pt off 4.5pt}]  (129.56,107.22) -- (194.89,61.22) ;

\draw    (22,144.01) -- (135.56,144.38) ;
\draw [shift={(137.56,144.39)}, rotate = 180.19] [fill={rgb, 255:red, 0; green, 0; blue, 0 }  ][line width=0.75]  [draw opacity=0] (8.93,-4.29) -- (0,0) -- (8.93,4.29) -- cycle    ;
\draw [shift={(20,144)}, rotate = 0.19] [fill={rgb, 255:red, 0; green, 0; blue, 0 }  ][line width=0.75]  [draw opacity=0] (8.93,-4.29) -- (0,0) -- (8.93,4.29) -- cycle    ;
\draw  [fill={rgb, 255:red, 255; green, 255; blue, 255 }  ,fill opacity=1 ] (523,90.67) .. controls (523,86.98) and (525.98,84) .. (529.67,84) .. controls (533.35,84) and (536.33,86.98) .. (536.33,90.67) .. controls (536.33,94.35) and (533.35,97.33) .. (529.67,97.33) .. controls (525.98,97.33) and (523,94.35) .. (523,90.67) -- cycle ;
\draw  [fill={rgb, 255:red, 255; green, 255; blue, 255 }  ,fill opacity=1 ] (523,108.67) .. controls (523,104.98) and (525.98,102) .. (529.67,102) .. controls (533.35,102) and (536.33,104.98) .. (536.33,108.67) .. controls (536.33,112.35) and (533.35,115.33) .. (529.67,115.33) .. controls (525.98,115.33) and (523,112.35) .. (523,108.67) -- cycle ;
\draw  [fill={rgb, 255:red, 255; green, 255; blue, 255 }  ,fill opacity=1 ] (523,126.67) .. controls (523,122.98) and (525.98,120) .. (529.67,120) .. controls (533.35,120) and (536.33,122.98) .. (536.33,126.67) .. controls (536.33,130.35) and (533.35,133.33) .. (529.67,133.33) .. controls (525.98,133.33) and (523,130.35) .. (523,126.67) -- cycle ;
\draw  [fill={rgb, 255:red, 255; green, 255; blue, 255 }  ,fill opacity=1 ] (523,145.67) .. controls (523,141.98) and (525.98,139) .. (529.67,139) .. controls (533.35,139) and (536.33,141.98) .. (536.33,145.67) .. controls (536.33,149.35) and (533.35,152.33) .. (529.67,152.33) .. controls (525.98,152.33) and (523,149.35) .. (523,145.67) -- cycle ;
\draw  [fill={rgb, 255:red, 255; green, 255; blue, 255 }  ,fill opacity=1 ] (523,163.67) .. controls (523,159.98) and (525.98,157) .. (529.67,157) .. controls (533.35,157) and (536.33,159.98) .. (536.33,163.67) .. controls (536.33,167.35) and (533.35,170.33) .. (529.67,170.33) .. controls (525.98,170.33) and (523,167.35) .. (523,163.67) -- cycle ;
\draw    (210,56) -- (210.33,209.33) ;

\draw    (225,56) -- (225.33,209.33) ;

\draw    (239,56) -- (239.33,209.33) ;

\draw    (253,55) -- (253.33,208.33) ;

\draw    (196,182) -- (268.44,182.33) ;

\draw    (195.33,168.67) -- (267.78,169) ;

\draw    (195.33,154.67) -- (267.78,155) ;

\draw    (195.33,140) -- (267.78,140.33) ;

\draw    (195.33,126.67) -- (267.78,127) ;

\draw    (195.33,112.67) -- (267.78,113) ;

\draw    (195.33,98.67) -- (267.78,99) ;

\draw    (196,84.67) -- (268.44,85) ;

\draw    (195.33,69.33) -- (267.78,69.67) ;

\draw  [fill={rgb, 255:red, 211; green, 211; blue, 211 }  ,fill opacity=1 ] (317.54,107.31) -- (317.23,178.42) -- (302.9,178.36) -- (303.21,107.25) -- cycle ;
\draw    (317.48,121.31) -- (303.93,121.03) ;

\draw    (316.75,135.97) -- (303.2,135.69) ;

\draw    (316.69,150.64) -- (303.13,150.36) ;

\draw    (316.63,164.64) -- (303.07,164.36) ;

\draw    (491.11,84.22) -- (475.78,84.22) ;

\draw  [dash pattern={on 0.84pt off 2.51pt}]  (490.67,77.42) -- (523,90.67) ;

\draw  [dash pattern={on 0.84pt off 2.51pt}]  (490.67,91.42) -- (523,108.67) ;

\draw  [dash pattern={on 0.84pt off 2.51pt}]  (491.33,104.75) -- (523,90.67) ;

\draw  [dash pattern={on 0.84pt off 2.51pt}]  (492,174.75) -- (523,90.67) ;

\draw  [dash pattern={on 0.84pt off 2.51pt}]  (492,174.75) -- (523,163.67) ;

\draw  [dash pattern={on 0.84pt off 2.51pt}]  (490.67,146.08) -- (523,163.67) ;

\draw  [dash pattern={on 0.84pt off 2.51pt}]  (490.67,146.08) -- (523,145.67) ;

\draw  [dash pattern={on 0.84pt off 2.51pt}]  (491.33,104.75) -- (523,145.67) ;

\draw  [dash pattern={on 0.84pt off 2.51pt}]  (490.67,132.42) -- (523,126.67) ;

\draw  [dash pattern={on 0.84pt off 2.51pt}]  (490.67,132.42) -- (523,108.67) ;

\draw  [dash pattern={on 0.84pt off 2.51pt}]  (490.67,132.42) -- (523,163.67) ;

\draw  [fill={rgb, 255:red, 255; green, 255; blue, 255 }  ,fill opacity=1 ] (574.33,126.67) .. controls (574.33,122.98) and (577.32,120) .. (581,120) .. controls (584.68,120) and (587.67,122.98) .. (587.67,126.67) .. controls (587.67,130.35) and (584.68,133.33) .. (581,133.33) .. controls (577.32,133.33) and (574.33,130.35) .. (574.33,126.67) -- cycle ;
\draw  [dash pattern={on 0.84pt off 2.51pt}]  (536.33,90.67) -- (574.33,126.67) ;

\draw  [dash pattern={on 0.84pt off 2.51pt}]  (536.33,108.67) -- (574.33,126.67) ;

\draw  [dash pattern={on 0.84pt off 2.51pt}]  (536.33,163.67) -- (574.33,126.67) ;

\draw  [dash pattern={on 0.84pt off 2.51pt}]  (536.33,145.67) -- (574.33,126.67) ;

\draw  [dash pattern={on 0.84pt off 2.51pt}]  (536.33,126.67) -- (574.33,126.67) ;

\draw    (587.67,126.67) -- (606.11,126.79) ;
\draw [shift={(608.11,126.81)}, rotate = 180.39] [fill={rgb, 255:red, 0; green, 0; blue, 0 }  ][line width=0.75]  [draw opacity=0] (10.72,-5.15) -- (0,0) -- (10.72,5.15) -- (7.12,0) -- cycle    ;

\draw    (491.11,99.22) -- (475.78,99.22) ;

\draw    (491.11,112.89) -- (475.78,112.89) ;

\draw    (490.78,125.89) -- (475.44,125.89) ;

\draw    (491.11,139.56) -- (475.78,139.56) ;

\draw    (491.11,154.56) -- (475.78,154.56) ;

\draw    (491.78,168.22) -- (476.44,168.22) ;

\draw (169,128.33) node  [align=left] {$\displaystyle V$};
\draw (237,25.67) node  [align=left] {$\displaystyle q$};
\draw (320.67,59.67) node  [align=left] {$\displaystyle S$};
\draw (422.5,126.42) node  [align=left] {GRU};
\draw (231,152) node  [align=left] {$ $\\\\\\};
\draw (620.08,125.75) node  [align=left] {$\displaystyle \hat{y}_{t}$};
\draw (28.33,120) node  [align=left] {1};
\draw (50,120) node  [align=left] {10};
\draw (71,120) node  [align=left] {2};
\draw (91,120) node  [align=left] {4};
\draw (110,120) node  [align=left] {0};
\draw (130,120) node  [align=left] {0};
\draw (30,78) node [scale=0.9] [align=left] {{\small it}};
\draw (49,78) node [scale=0.9] [align=left] {{\small will}};
\draw (70,78) node [scale=0.9] [align=left] {{\small be}};
\draw (92.67,78) node [scale=0.9] [align=left] {{\small cold}};
\draw (230,230) node  [align=left] {{\small Embedding matrix}};
\draw (338.67,230) node  [align=left] {{\small  \ Sentence as}\\{\small seq. of vectors}};
\draw (481,229) node  [align=left] {{\small  GRU}\\{\small output}};
\draw (596,230.67) node  [align=left] {{\small Forecast}\\{\small  \ day $\displaystyle t$}};
\draw (80.67,229) node  [align=left] {{\small Report of day $\displaystyle t$}};
\draw (76.33,155.67) node  [align=left] {$\displaystyle S$};
\draw (530,187) node  [align=left] {{\small Dense}};
\draw (310.67,183.67) node  [align=left] {{\scriptsize 1}};
\draw (323.33,174.67) node  [align=left] {{\scriptsize 10}};
\draw (334.33,164.33) node  [align=left] {{\scriptsize 2}};
\draw (342.67,154.67) node  [align=left] {{\scriptsize 4}};
\draw (352.67,146) node  [align=left] {{\scriptsize 0}};
\draw (361,136) node  [align=left] {{\scriptsize 0}};

\end{tikzpicture}

    \caption{Structure of our RNN. Dropout and batch normalization are not represented.}
    \label{Fig:RNN_structure}
\end{figure}

As for the word embedding, recurrent or convolutional neural networks (respectively RNN and CNN) were used with them. MLPs are not used, for they would require to concatenate all the vector representations of a sentence together beforehand and result in a network with too many parameters to be trained correctly with our number of available documents. Recall that we decided to train our own vector representation of words instead of using an already available one. In order to obtain the embedding, the texts are first converted into a sequence of integers: each word is given a number ranging from 1 to $V$, where $V$ is the vocabulary size (0 is used for padding or unknown words in the test set). One must then calculate the maximum sequence length $S$, and sentences of length shorter than $S$ are then padded by zeros. During the training process of the network, for each word a $q$ dimensional real-valued vector representation is calculated simultaneously to the rest of the weights of the network. Ergo a sentence of $S$ words is translated into a sequence of $S$ $q$-sized vectors, which is then fed into a recurrent neural unit. For both languages, $q=20$ seemed to yield the best results. In the case of recurrent units two main possibilities arise, with LSTM (Long Short-Term Memory) \cite{hochreiter1997long} and GRU (Gated Recurrent Unit) \cite{cho2014learning}. After a few initial trials, no significant performance differences were noticed between the two types of cells. Therefore GRU were systematically used for recurrent networks, since their lower amount of parameters makes them easier to train and reduces overfitting. The output of the recurrent unit is afterwards linked to a fully connected (also referred as dense) layer, leading to the final forecast as output. The rectified linear unit (ReLU) activation in dense layers systematically gave the best results, except on the output layer where we used a sigmoid one considering the time series' normalization. In order to tone down overfitting, dropout layers \cite{srivastava2014dropout} with probabilities of 0.25 or 0.33 are set in between the layers. Batch normalization \cite{batchnorm} is also used before the GRU since it stabilized training and improved performance. Figure \ref{Fig:RNN_structure} represents the architecture of our RNN. 

The word embedding matrix is therefore learnt jointly with the rest of the parameters of the neural network by minimization of the quadratic loss with respect to the true electricity demand. Note that while above we described the case of the RNN, the same procedure is considered for the case of the CNN, with only the recurrent layers replaced by a combination of 1D convolution and pooling ones. As for the optimization algorithms of the neural networks, traditional stochastic gradient descent with momentum or ADAM \cite{kingma2014adam} together with a quadratic loss are used. All of the previously mentioned methods were coded with Python. The LASSO and RF were implemented using the library \texttt{Scikit Learn} \cite{scikit-learn}, while \texttt{Keras} \cite{chollet2015keras} was used for the neural networks.

\subsection{Hyperparameter Tuning}

While most parameters are trained during the learning optimization process, all methods still involve a certain number of hyperparameters that must be manually set by the user. For instance for random forests it can correspond to the maximum depth of the trees or the fraction of features used at each split step, while for neural networks it can be the number of layers, neurons, the embedding dimension or the activation functions used. This is why the data is split into three sets:
\begin{itemize}
    \item The training set, using all data available up to the 31\textsuperscript{st} of December 2013 (2,557 days for France and 2,922 for the UK). It is used to learn the parameters of the algorithms through mathematical optimization.
    \item The years 2014 and 2015 serve as validation set (730 days). It is used to tune the hyperparameters of the different approaches.
    \item All the data from January the 1\textsuperscript{st} 2016 (974 days for France and 1,096 for the UK) is used as test set, on which the final results are presented.
\end{itemize}

Grid search is applied to find the best combination of values: for each hyperparameter, a range of values is defined, and all the possible combinations are successively tested. The one yielding the lowest RMSE (see section \ref{Sec:ExperimentalResults}) on the validation set is used for the final results on the test one. While relatively straightforward for RFs and the LASSO, the extreme number of possibilities for NNs and their extensive training time compelled us to limit the range of architectures possible. The hyperparameters are tuned per method and per country: ergo the hyperparameters of a given algorithm will be the same for the different time series of a country (e.g. the RNN architecture for temperature and load for France will be the same, but different from the UK one). Finally before application on the testing set, all the methods are re-trained from scratch using both the training and validation data. \\


\section{Experiments}
\label{Sec:ExperimentalResults}

The goal of our experiments is to quantify how close one can get using textual data only when compared to numerical data. However the inputs of the numerical benchmark should be hence comparable to the information contained in the weather reports. Considering they mainly contain calendar (day of the week and month) as well as temperature and wind information, the benchmark of comparison is a random forest trained on four features only: the time of the year (whose value is 0 on January the 1\textsuperscript{st} and 1 on December the 31\textsuperscript{st} with a linear growth in between), the day of the week, the national average temperature and wind speed. The metrics of evaluation are the Mean Absolute Percentage Error (MAPE), Root Mean Squared Error (RMSE), Mean Absolute Error (MAE) and the $R^2$ coefficient given by:

$$
    \begin{array}{lcl}
         \text{MAPE} & = & \displaystyle\dfrac{100}{T} \sum_{t=1}^T \left| \dfrac{y_t - \hat{y}_t}{y_t} \right|  \\
         \\
         \text{MAE} & = & \dfrac{1}{T} \displaystyle\sum_{t=1}^T \left| y_t - \hat{y}_t \right| \\
         \\
         \text{RMSE} & = & \sqrt{ \dfrac{1}{T} \displaystyle\sum_{t=1}^T \left( y_t - \hat{y}_t \right)^2 }\\
         \\
         R^2 & = & 1 - \dfrac{\sum_{t=1}^T (y_t - \hat{y}_t)^2}{\sum_{t=1}^T (y_t - \overline{y})^2}
    \end{array}
    \label{eq:error_metrics}
$$

\noindent where $T$ is the number of test samples, $y_t$ and $\hat{y}_t$ are respectively the ground truth and the prediction for the document of day $t$, and $\overline{y}$ is the empirical average of the time series over the test sample. A known problem with MAPE is that it unreasonably increases the error score for values close to 0. While for the load it isn't an issue at all, it can be for the meteorological time series. Therefore for the temperature, the MAPE is calculated only when the ground truth is above the 5\% empirical quantile. Although we aim at achieving the highest accuracy possible, we focus on the interpretability of our models as well.

\subsection{Feature selection}

Many words are obviously irrelevant to the time series in our texts. For instance the day of the week, while playing a significant role for the load demand, is useless for temperature or wind. Such words make the training harder and may decrease the accuracy of the prediction. Therefore a feature selection procedure similar to \cite{ben2007selection} is applied to select a subset of useful features for the different algorithms, and for each type of time series. Random forests are naturally able to calculate feature importance through the calculation of error increase in the out-of-bag (OOB) samples. Therefore the following process is applied to select a subset of $V^*$ relevant words to keep:

\begin{enumerate}
    \item A RF is trained on the whole training \& validation set. The OOB feature importance can thus be calculated. 
    \item The features are then successively added to the RF in decreasing order of feature importance.
    \item This process is repeated $B=10$ times to tone down the randomness. The number $V^*$ is then set to the number of features giving the highest median OOB $R^2$ value.
\end{enumerate}

\begin{figure}[H]
    \centering
    \includegraphics[scale=0.45]{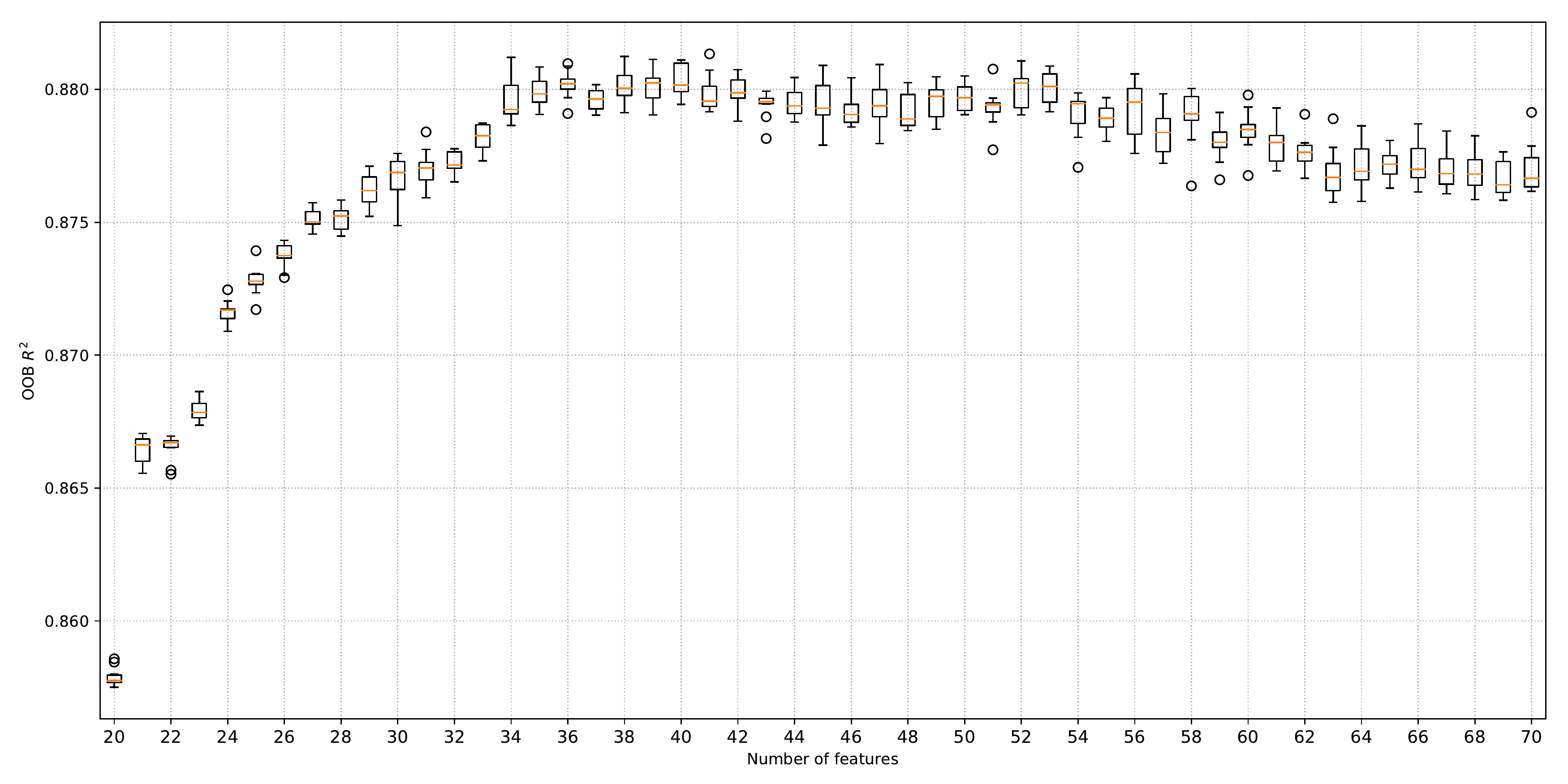}
    \caption{Evolution of the OOB $R^2$ during the selection procedure.}
    \label{Fig:Fr_OOB_featsel}
\end{figure}

The results of this procedure for the French data is represented in figure \ref{Fig:Fr_OOB_featsel}. The best median $R^2$ is achieved for $V^* = 52$, although one could argue that not much gain is obtained after $36$ words. The results are very similar for the UK data set, thus for the sake of simplicity the same value $V^* = 52$ is used. Note that the same subset of words is used for all the different forecasting models, which could be improved in further work using other selection criteria (e.g. mutual information, see \cite{mRMR}). An example of normalized feature importance is given in figure. \ref{Fig:RF_featimp}.

\subsection{Main results}

Note that most of the considered algorithms involve randomness during the training phase, with the subsampling in the RFs or the gradient descent in the NNs for instance. In order to tone it down and to increase the consistency of our results, the different models are run $B=10$ times. The results presented hereafter correspond to the average and standard-deviation on those runs. The RF model denoted as "sel" is the one with the reduced number of features, whereas the other RF uses the full vocabulary. We also considered an aggregated forecaster (abridged Agg), consisting of the average of the two best individual ones  in terms of RMSE. All the neural network methods have a reduced vocabulary size $V^*$. The results for the French and UK data are respectively given by tables \ref{Tab:err_FR} and \ref{Tab:err_UK}.

\begin{table}[H]
    \centering
    \begin{tabular}{|c|c|c|c|c|c|}
        \hline
        Method & Encoding & MAPE (\%) & RMSE (MW) & MAE (MW) & $R^2$ \\
        \hline
        Benchmark & - & $2.74 \pm 0.01$ & $2152.86\pm 5.51$ & $1533.65\pm 3.38$ & $0.96\pm 0.00$ \\
        \hline
        RF & \multirow{4}{*}{TF-IDF} & $5.40 \pm 0.02$ & $4386.67 \pm 15.67$ & $3052.75\pm 9.07$ & $0.85 \pm 0.00$ \\
        RF (sel) & & $5.28 \pm 0.01$ & $4168.68 \pm 12.13$ & $2977.76 \pm 6.28$ & $0.86 \pm 0.00$ \\
        MLP &  & $5.09 \pm 0.05$ & $3720.94 \pm 27.29 $ & $2795.61 \pm 25.78$ & $0.89 \pm 0.00$ \\
        LASSO & & $ 6.81 \pm 0.00$ & $4829.99 \pm 0.00$ & $3717.84 \pm 0.00$ & $0.82 \pm 0.00$ \\
        \hline
        CNN & \multirow{2}{*}{Embedding} & $6.18 \pm 0.34$ & $4493.08 \pm 171.40$ & $3359.61 \pm 165.39$ & $0.84 \pm 0.01$ \\
        RNN & & $4.83 \pm 0.11$ & $3577.04 \pm 64.46$ & $2672.12 \pm 52.41$ & $0.90 \pm 0.00$ \\
        \hline
        Agg (RNN + MLP) & Both & $4.66 \pm 0.04$ & $3470.53 \pm 22.25$ & $2581.94\pm 21.67$ & $0.91 \pm 0.00$ \\ 
        \hline
    \end{tabular}
    \caption{Forecast errors on the net load for the French Dataset.}
    \label{Tab:err_FR}
\end{table}

\begin{table}[H]
    \centering
    \begin{tabular}{|c|c|c|c|c|c|}
        \hline
        Method & Encoding & MAPE (\%) & RMSE (MW) & MAE (MW) & $R^2$ \\
        \hline
        Benchmark & - & $2.28 \pm 0.00$ & $1013.83 \pm 2.16$ & $754.53 \pm 1.45$ & $0.94\pm 0.00$ \\
        \hline
        RF & \multirow{4}{*}{TF-IDF} & $3.19 \pm 0.01$ & $ 1544.81\pm 3.34$ & $1058.16 \pm 2.66$ & $0.85 \pm 0.00$ \\
        RF (sel) & & $3.22 \pm 0.01$ & $1545.04 \pm 3.90$ & $1072.00\pm3.19$ & $0.85 \pm 0.00$ \\
        MLP &  & $3.68 \pm 0.13$ & $1651.10 \pm 36.46$ & $1209.16 \pm 39.34 $ & $0.83\pm 0.01$ \\
        LASSO & & $4.70\pm 0.00$  & $2257.32\pm 0.00$  & $1519.35 \pm 0.00$  & $0.69 \pm 0.00$ \\
        \hline
        CNN & \multirow{2}{*}{Embedding} & $4.23\pm 0.24$ & $1910.73\pm 66.39 $ & $1387.22\pm 69.37$ & $0.78\pm 0.02$ \\
        RNN & & $3.55 \pm 0.08$ & $1695.17 \pm 40.82$ & $1182.31\pm 26.41$ & $0.82\pm 0.01$ \\
        \hline
        Agg (RF + RF Sel) & TF-IDF & $ 3.18\pm 0.01$ & $1532.75\pm 2.66 $ & $1056.55\pm 2.42$ & $ 0.86\pm 0.00$ \\ 
        \hline
    \end{tabular}
    \caption{Forecast errors on the net load for the British Dataset.}
    \label{Tab:err_UK}
\end{table}

Our empirical results show that for the electricity consumption prediction task, the order of magnitude of the relative error is around 5\%, independently of the language, encoding and machine learning method, thus proving the intrinsic value of the information contained in the textual documents for this time series. As expected, all text based methods perform poorer than when using explicitly numerical input features. Indeed, despite containing relevant information, the text is always more fuzzy and less precise than an explicit value for the temperature or the time of the year for instance. Again the aim of this work is not to beat traditional methods with text, but quantifying how close one can come to traditional approaches when using text exclusively. As such achieving less than 5\% of MAPE was nonetheless deemed impressive by expert electricity forecasters. Feature selection brings significant improvement in the French case, although it does not yield any improvement in the English one. The reason for this is currently unknown. Nevertheless the feature selection procedure also helps the NNs by dramatically reducing the vocabulary size, and without it the training of the networks was bound to fail. While the errors accross methods are roughly comparable and highlight the valuable information contained within the reports, the best method nonetheless fluctuates between languages. Indeed in the French case there is a hegemony of the NNs, with the embedding RNN edging the MLP TF-IDF one. However for the UK data set the RFs yield significantly better results on the test set than the NNs. This inversion of performance of the algorithms is possibly due to a change in the way the reports were written by the Met Office after August 2017, since the results of the MLP and RNN on the validation set (not shown here) were satisfactory and better than both RFs. For the two languages both the CNN and the LASSO yielded poor results. For the former, it is because despite grid search no satisfactory architecture was found, whereas the latter is a linear approach and was used more for interpretation purposes than strong performance. Finally the naive aggregation of the two best experts always yields improvement, especially for the French case where the two different encodings are combined. This emphasises the specificity of the two representations leading to different types of errors. An example of comparison between ground truth and forecast for the case of electricity consumption is given for the French language with fig. \ref{Fig:overlap_FR}, while another for temperature may be found in the appendix \ref{Fig:overlap_FR_Temp}. The sudden "spikes" in the forecast are due to the presence of winter related words in a summer report. This is the case when used in comparisons, such as \textit{"The flood will be as severe as in January"} in a June report and is a limit of our approach. Finally, the usual residual $\hat{\varepsilon}_t = y_t - \hat{y}_t$ analyses procedures were applied: Kolmogorov normality test, QQplots comparaison to gaussian quantiles, residual/fit comparison... While not thoroughly gaussian, the residuals were close to normality nonetheless and displayed satisfactory properties such as being generally independent from the fitted and ground truth values. Excerpts of this analysis for France are given in figure \ref{Fig:FR_residuals_analysis} of the appendix. The results for the temperature and wind series are given in appendix. Considering that they have a more stochastic behavior and are thus more difficult to predict, the order of magnitude of the errors differ (the MAPE being around 15\% for temperature for instance) but globally the same observations can be made. \\

\begin{table}[H]
    \centering
    \begin{tabular}{|c|c|c|c|c|c|c|}
        \hline
        Country & Predicted series & Method + Encoding & MAPE (\%) & RMSE & MAE & $R^2$\\
        \hline
        \multirow{3}{*}{France} & Load & RNN + Embedding & $4.83$ & $3577.04$ & $2672.12$ & $0.90$ \\
                                & Temperature & RNN + Embedding & $14.60$ & $1.95$ & $1.53$ & $0.91$  \\
                                & Wind & RNN + Embedding & $18.11$ & $0.68$ & $0.53$ & $0.51$ \\
        \hline
        \multirow{3}{*}{UK} & Load & RF + TF-IDF & 3.19 & 1544.81 & 1058.16 & 0.85 \\
                            & Temperature & RNN + Embedding & 15.80 & 1.58 & 1.24 & 0.89  \\
                            & Wind & LASSO + TF-IDF & 19.76 & 0.99 & 0.78 & 0.53 \\
        \hline
    \end{tabular}
    \caption{Best (individual, in terms of RMSE) result for each of the considered time series.}
    \label{Tab:best_pred_by_feature}
\end{table}

\begin{figure}[H]
    \centering
    \includegraphics[scale=0.39]{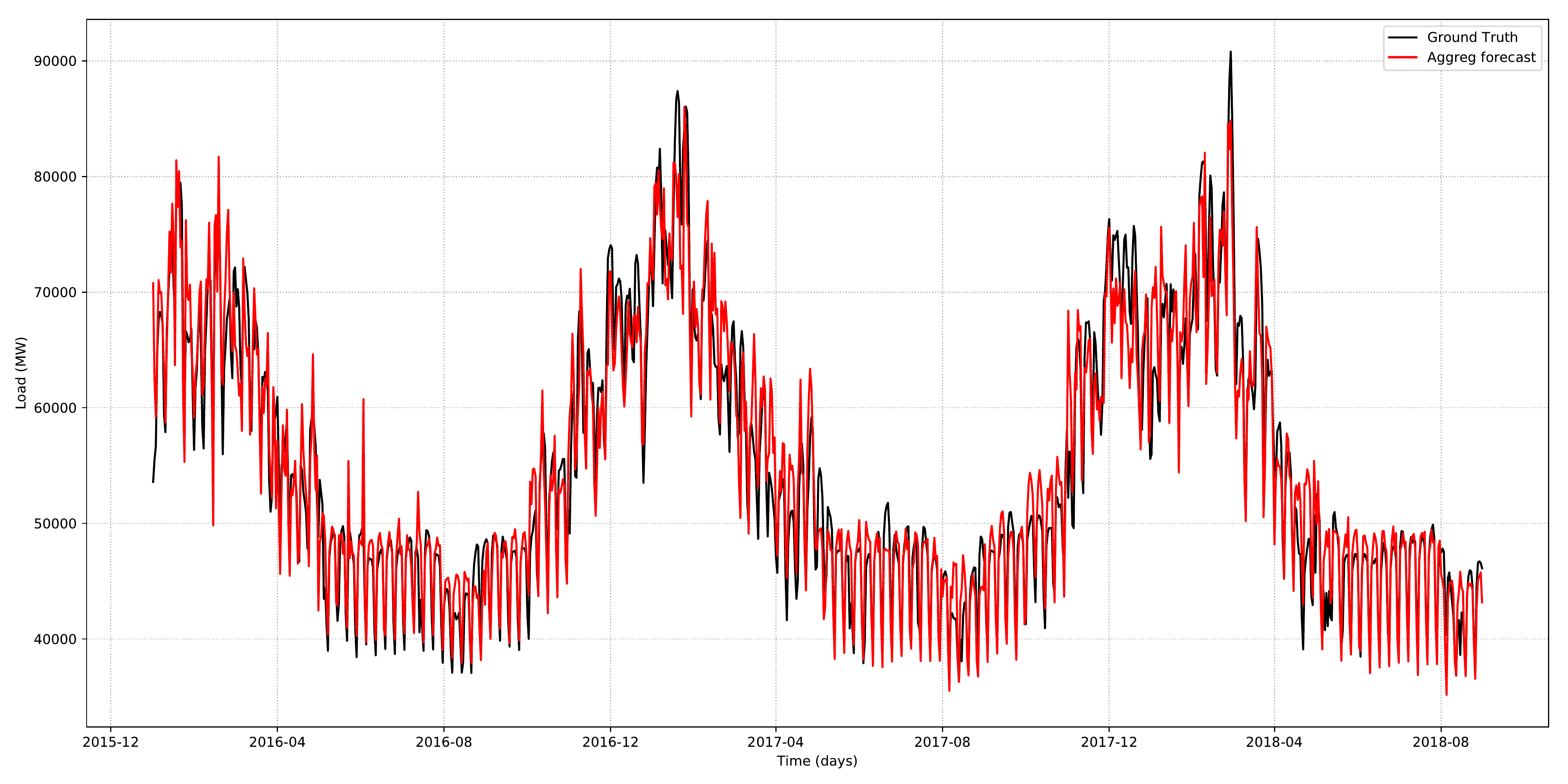}
    \caption{Overlapping of prediction and real load (France)}
    \label{Fig:overlap_FR}
\end{figure}

\subsection{Interpretability of the models}

While accuracy is the most relevant metric to assess forecasts, interpretability of the models is of paramount importance, especially in the field of professional electricity load forecasting and considering the novelty of our work. Therefore in this section we discuss the properties of the RF and LASSO models using the TF-IDF encoding scheme, as well as the RNN word embedding.

\subsubsection{TF-IDF representation}

\begin{figure}[H]
    \centering

    \begin{subfigure}[b]{0.49\textwidth}
        \includegraphics[width=\textwidth]{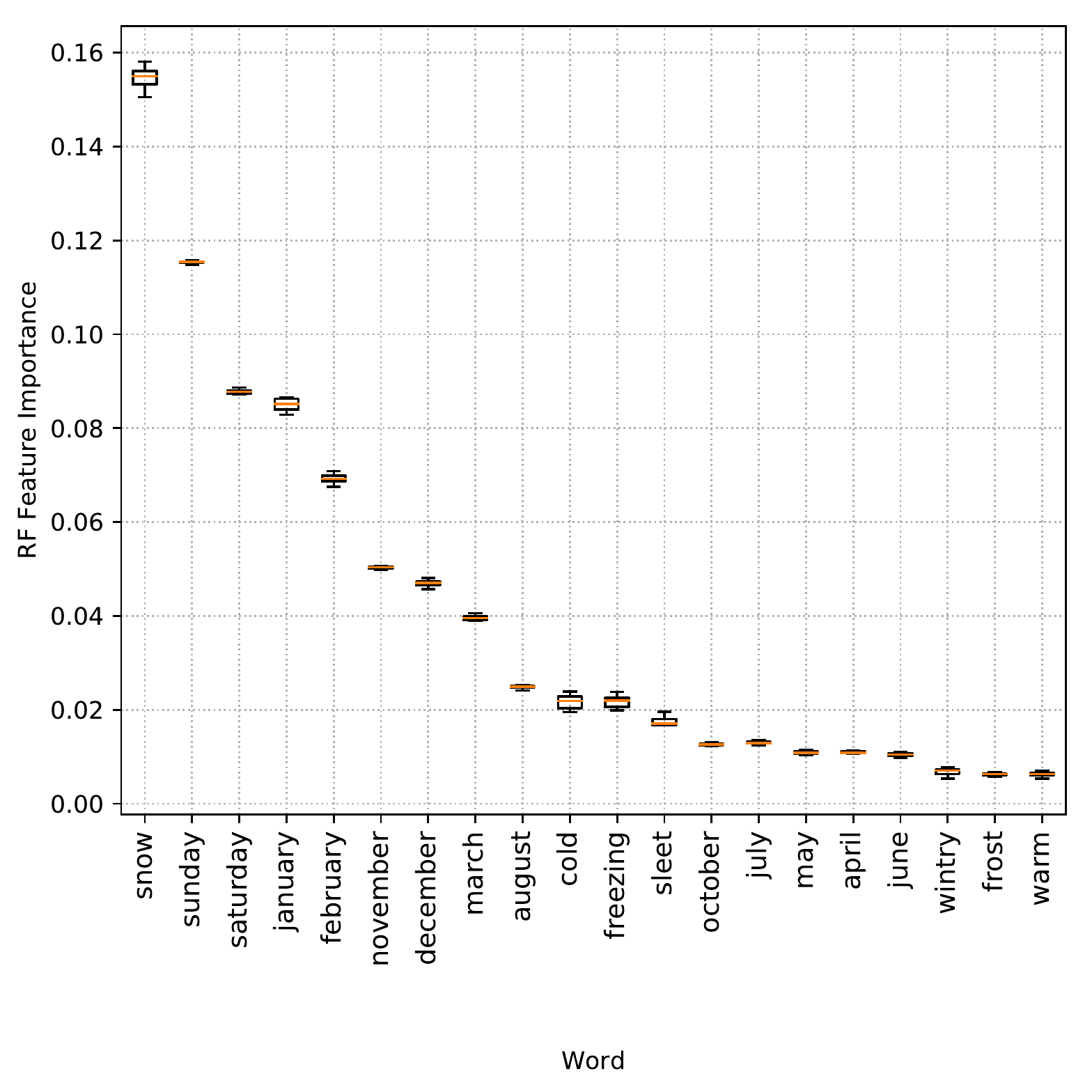}
        \caption{UK}
    \end{subfigure} \hspace{-0.2cm} 
    \begin{subfigure}[b]{0.49\textwidth}
        \includegraphics[width=\textwidth]{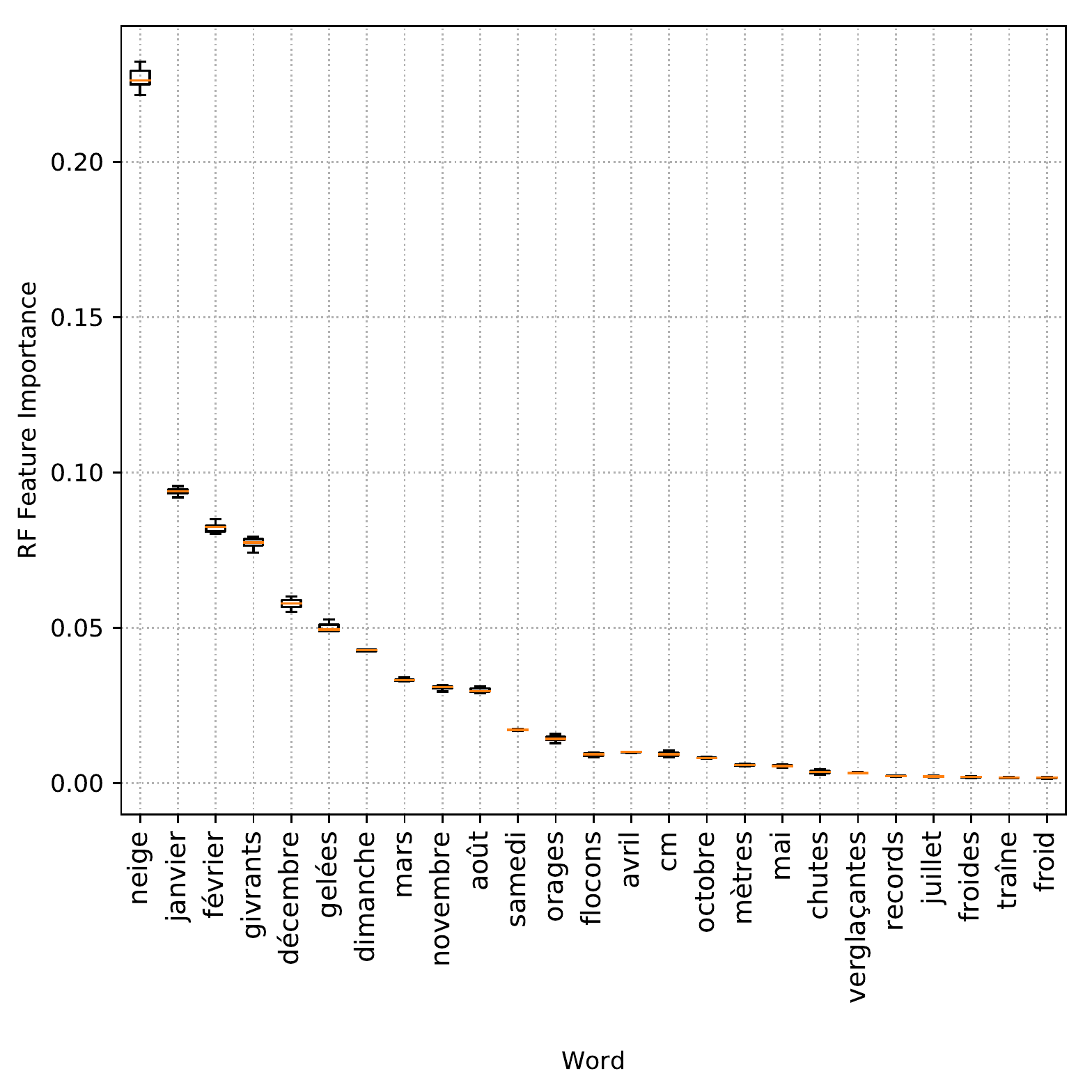}
        \caption{France}
    \end{subfigure}
    
    \caption{RF feature importance over the $B=10$ runs.} \label{Fig:RF_featimp}
\end{figure}

One significant advantage of the TF-IDF encoding when combined with random forests or the LASSO is that it is possible to interpret the behavior of the models. For instance, figure \ref{Fig:RF_featimp} represents the 20 most important features (in the RF OOB sense) for both data sets when regressing over electricity demand data. As one can see, the random forest naturally extracts calendar information contained in the weather reports, since months or week-end days are among the most important ones. For the former, this is due to the periodic behavior of electricity consumption, which is higher in winter and lower in summer. This is also why characteristic phenomena of summer and winter, such as "thunderstorms", "snow" or "freezing" also have a high feature importance. The fact that August has a much more important role than July also concurs with expert knowledge, especially for France: indeed it is the month when most people go on vacations, and thus when the load drops the most. As for the week-end names, it is due to the significantly different consumer behavior during Saturdays and especially Sundays when most of the businesses are closed and people are usually at home. Therefore the relevant words selected by the random forest are almost all in agreement with expert knowledge.

\begin{figure}[H]
    \centering
    \includegraphics[scale=0.45]{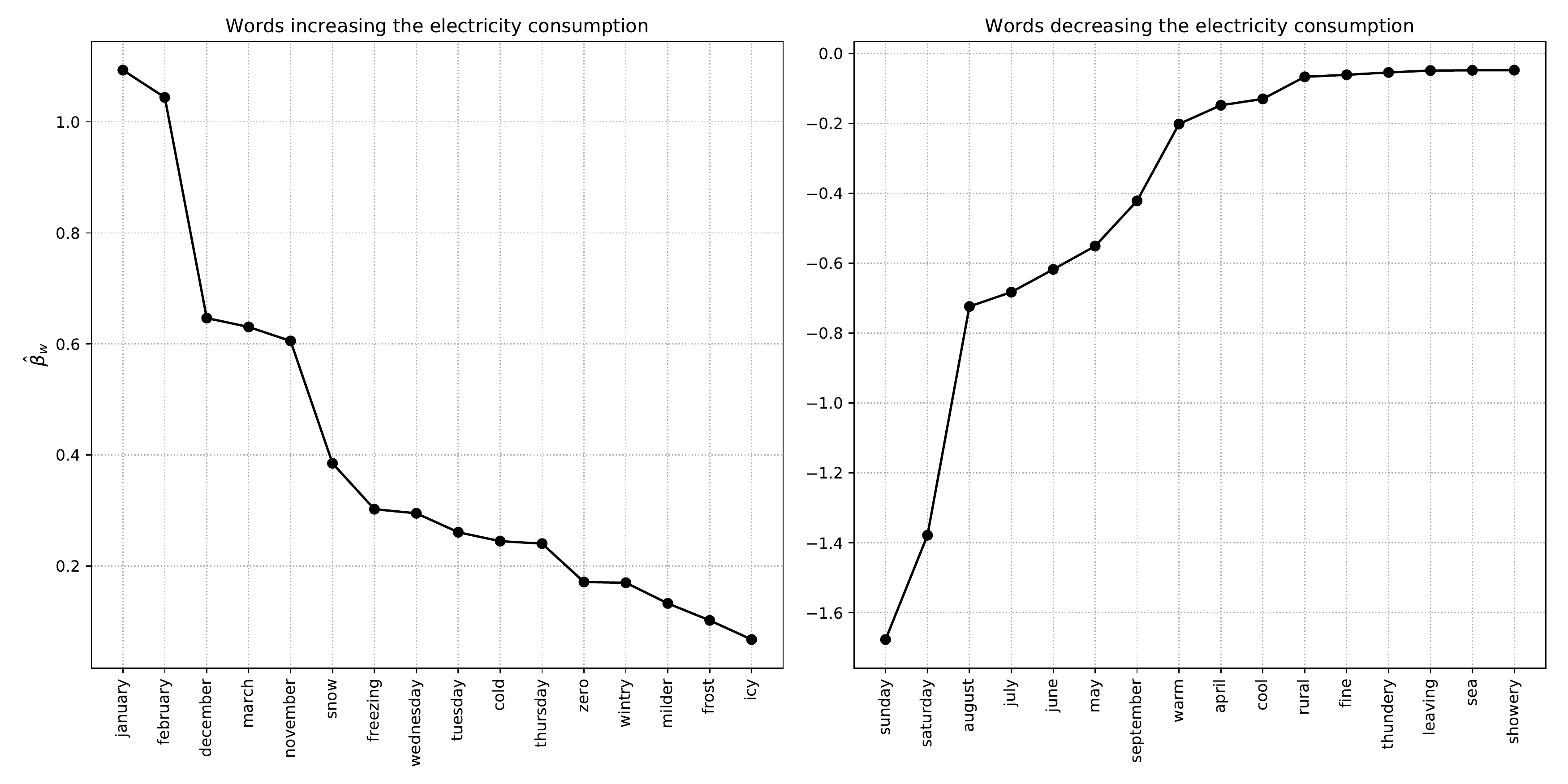}
    \caption{Coefficients $\beta_w$ in the british load LASSO regression.}
    \label{Fig:LASSO_coeffs_UK}
\end{figure}

We also performed the analysis of the relevant words for the LASSO. In order to do that, we examined the words $w$ with the largest associated coefficients $\beta_w$ (in absolute value) in the regression. Since the TF-IDF matrix has positive coefficients, it is possible to interpret the sign of the coefficient $\beta_w$ as its impact on the time series. For instance if $\beta_w > 0$ then the presence of the word $w$ causes a rise the time series (respectively if $\beta_w < 0$, it entails a decline). The results are plotted fig. \ref{Fig:LASSO_coeffs_UK} for the the UK. As one can see, the winter related words have positive coefficients, and thus increase the load demand as expected whereas the summer related ones decrease it. The value of the coefficients also reflects the impact on the load demand. For example January and February have the highest and very similar values, which concurs with the similarity between the months. Sunday has a much more negative coefficient than Saturday, since the demand significantly drops during the last day of the week. The important words also globally match between the LASSO and the RF, which is a proof of the consistency of our results (this is further explored afterwards in figure \ref{Fig:Venn}). Although not presented here, the results are almost identical for the French load, with approximately the same order of relevancy. The important words logically vary in function of the considered time series, but are always coherent. For instance for the wind one, terms such as "gales", "windy" or "strong" have the highest positive coefficients, as seen in the appendix figure \ref{Fig:LASSO_coeffs_wind_UK}. Those results show that a text based approach not only extracts the relevant information by itself, but it may eventually be used to understand which phenomena are relevant to explain the behavior of a time series, and to which extend.



\subsubsection{Vector embedding representation}
\label{Sec:embedding}

Word vector embeddings such as Word2Vec and GloVe are known for their vectorial properties translating linguistic ones. However considering the objective function of our problem, there was no obvious reason for such attributes to appear in our own. Nevertheless for both languages we conducted an analysis of the geometric properties of our embedding matrix. We investigated the distances between word vectors, the relevant metric being the cosine distance given by:

$$
    \text{dcos}(\overrightarrow{w_1},\overrightarrow{w_2}) = 1 - \dfrac{\overrightarrow{w_1} \cdot \overrightarrow{w_2}}{\norm{\overrightarrow{w_1}}_2  \norm{\overrightarrow{w_2}}_2}
    \label{eq:dist_cosinus}
$$

\noindent where $\overrightarrow{w_1}$ and $\overrightarrow{w_2}$ are given word vectors. Thus a cosine distance lower than 1 means similarity between word vectors, whereas a greater than 1 corresponds to opposition.

\begin{figure}[H]
    \centering
    \begin{subfigure}[b]{0.49\textwidth}
        \includegraphics[width=\textwidth]{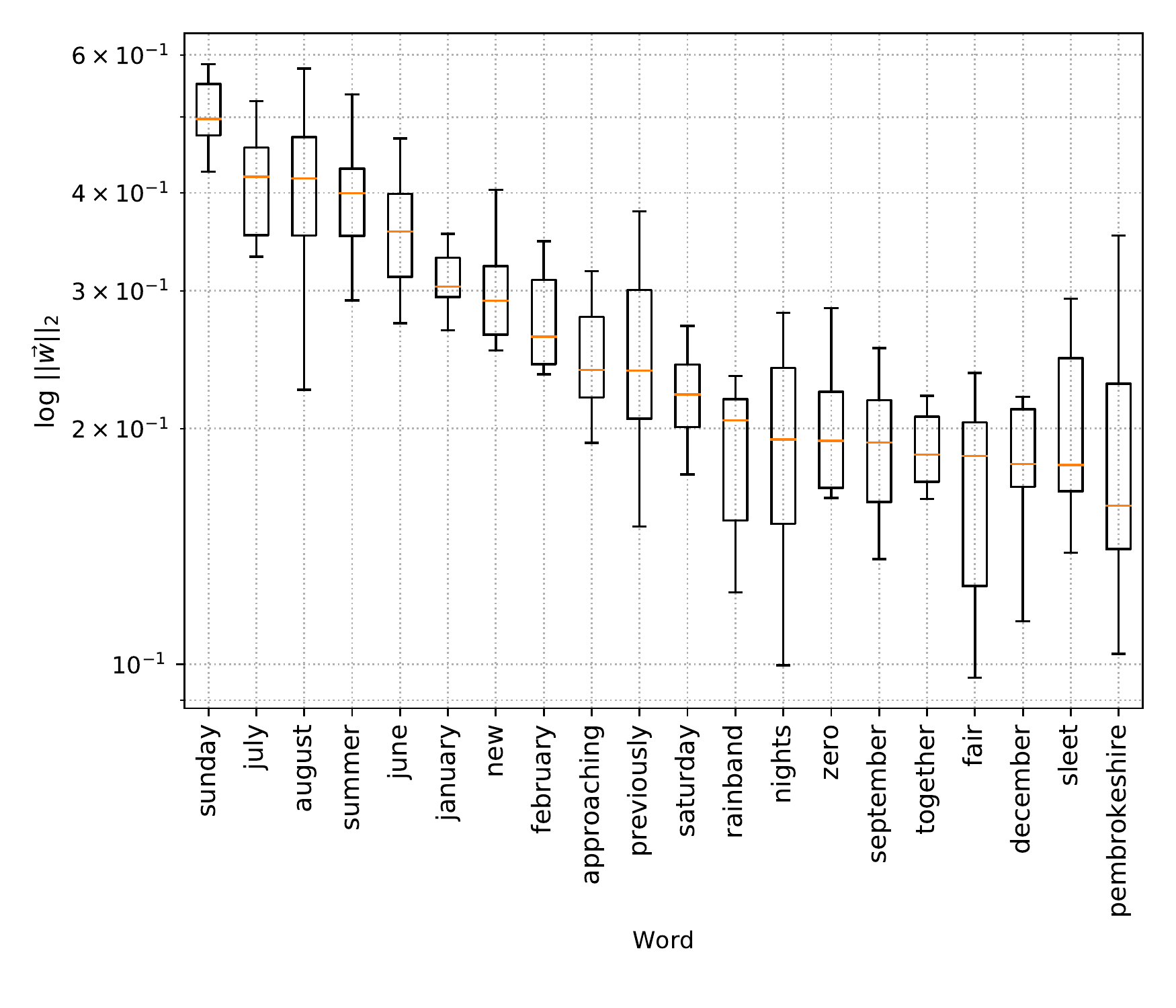}
        \caption{UK}
    \end{subfigure}
    \hspace{-0.2cm} 
    \begin{subfigure}[b]{0.49\textwidth}
        \includegraphics[width=\textwidth]{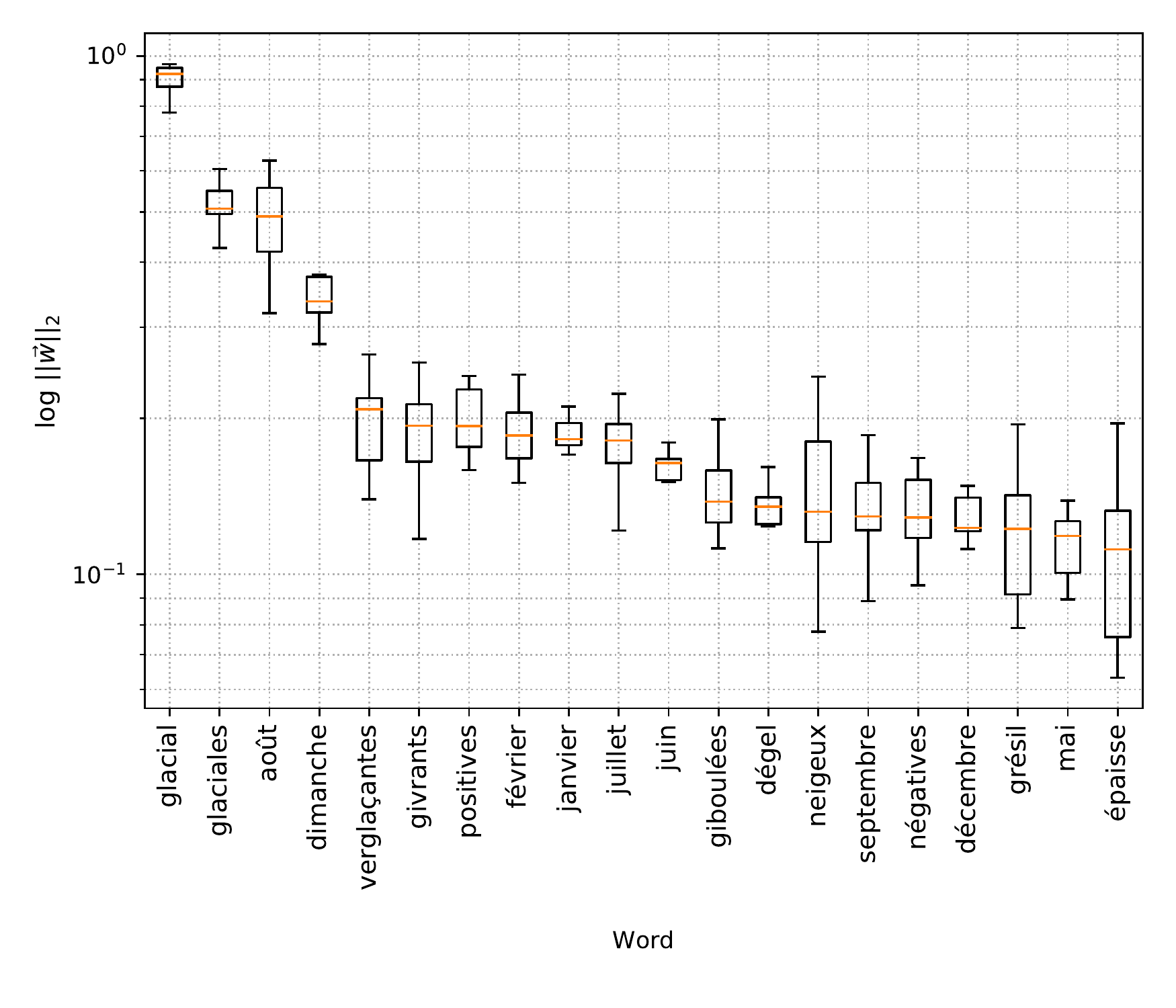}
        \caption{France}
    \end{subfigure}
    \caption{Word vector log-norm over $B=10$.} \label{Fig:embedding_norm}
\end{figure}

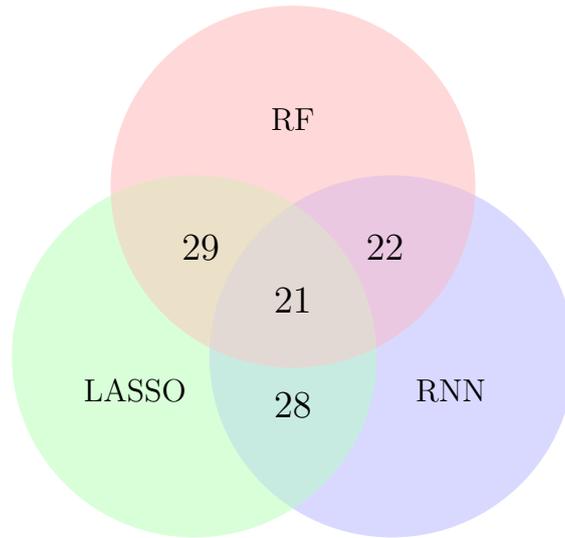
\begin{figure}[H]
    \centering
    \begin{tikzpicture}
      \begin{scope}[blend group = soft light]
        \fill[red!30!white,opacity=0.5]   ( 90:1.5) circle (2.4);
        \fill[green!30!white,opacity=0.5] (210:1.5) circle (2.4);
        \fill[blue!30!white,opacity=0.5]  (330:1.5) circle (2.4);
      \end{scope}
      \node [font=\large] at ( 90:2.4)    {RF};
      \node [font=\large] at ( 210:2.4)   {LASSO};
      \node [font=\large] at ( 330:2.4)   {RNN};
      \node [font=\Large] {21};
      \node [font=\Large] at ( 30:1.4) {22};
      \node [font=\Large] at ( 150:1.4) {29};
      \node [font=\Large] at ( 270:1.4) {28};
    \end{tikzpicture}
    \caption{Venn diagram of common words among the top 50 ones for each selection procedure (France).}
    \label{Fig:Venn}
\end{figure}

The initial analyses of the embedding matrices for both the UK and France revealed that in general, words were grouped by context or influence on the electricity consumption. For instance, we observed that winter words were together and far away from summer ones. Week days were grouped as well and far from week-end days. However considering the vocabulary was reduced to $V^* = 52$ words, those results lacked of consistency. Therefore for both languages we decided to re-train the RNNs using the same architecture, but with a larger vocabulary of the $V=300$ most relevant words (still in the RF sense) and on all the available data (i.e. everything is used as training) to compensate for the increased size of the vocabulary. We then calculated the distance of a few prominent words to the others. The analysis of the average cosine distance over $B=10$ runs for three major words is given by tables  \ref{Tab:cosine_words_UK1} and \ref{Tab:cosine_words_FR1}, and three other examples are given in the appendix tables \ref{Tab:cosine_words_UK2} and \ref{Tab:cosine_words_FR2}. The first row corresponds to the reference word vector $\overrightarrow{w_1}$ used to calculate the distance from (thus the distance is always zero), while the following ones are the 9 closest to it. The two last rows correspond to words we deemed important to check the distance with (an antagonistic one or relevant one not in the top 9 for instance). \\
The results of the experiments are very similar for both languages again. Indeed, the words are globally embedded in the vector space by topic: winter related words such as "January" ("janvier"), "February" ("février"), "snow" ("neige"), "freezing" ("glacial") are close to each other and almost opposite to summer related ones such as "July" ("juillet"), "August" ("août"), "hot" ("chaud"). For both cases the week days Monday ("lundi") to Friday ("vendredi") are grouped very closely to each other, while significantly separated from the week-end ones "Saturday" ("samedi") and "Sunday" ("dimanche"). Despite these observations, a few seemingly unrelated words enter the lists of top 10, especially for the English case (such as "pressure" or "dusk" for "February"). In fact the French language embedding seems of better quality, which is perhaps linked to the longer length of the French reports in average. This issue could probably be addressed with more data. Another observation made is that the importance of a word $w$ seems related to its euclidean norm in the embedding space $\norm{\overrightarrow{w}}_2$. For both languages the list of the 20 words with the largest norm is given fig. \ref{Fig:embedding_norm}. As one can see, it globally matches the selected ones from the RF or the LASSO (especially for the French language), although the order is quite different. This is further supported by the Venn diagram of common words among the top 50 ones for each word selection method represented in figure \ref{Fig:Venn} for France. Therefore this observation could also be used as feature selection procedure for the RNN or CNN in further work.

\begin{table}[H]
\parbox{.3\linewidth}{
\centering
\begin{tabular}{|c|c|}
\hline
Word & dcos $\pm \, \sigma$ \\
\hline
february &     0.000 $\pm$  0.000 \\
january  &     0.099 $\pm$  0.029 \\
zero     &     0.287 $\pm$  0.071 \\
milder   &     0.359 $\pm$  0.099 \\
coldest  &     0.410 $\pm$  0.151 \\
pressure &     0.430 $\pm$  0.136 \\
dusk     &     0.435 $\pm$  0.085 \\
november &     0.445 $\pm$  0.050 \\
isles    &     0.461 $\pm$  0.148 \\
freezing &     0.465 $\pm$  0.133 \\
\hline
july     &    1.595 $\pm$ 0.128 \\
august   &    1.596 $\pm$ 0.103 \\
\hline
\end{tabular}
\caption*{February}
\label{Tab:February_UK}
}
\hfill
\parbox{.3\linewidth}{
\centering
\begin{tabular}{|c|c|}
\hline
Word &  dcos $\pm \, \sigma$ \\
\hline
snow          &     0.000 $\pm$  0.000 \\
freezing      &     0.255 $\pm$  0.063 \\
zero          &     0.275 $\pm$  0.058 \\
milder        &     0.310 $\pm$  0.066 \\
accumulations &     0.344 $\pm$  0.107 \\
icy           &     0.351 $\pm$ 0.128 \\
coldest       &     0.384 $\pm$  0.122 \\
cold          &     0.387 $\pm$  0.100 \\
wintry        &     0.440 $\pm$  0.098 \\
january       &     0.441 $\pm$  0.132 \\
\hline
august        &     1.262 $\pm$  0.158 \\
hot         &     1.137 $\pm$  0.122 \\
\hline
\end{tabular}
\caption*{Snow}
\label{Tab:Snow_UK}
}
\hfill
\parbox{.3\linewidth}{
\centering
\begin{tabular}{|c|c|}
\hline
Word & dcos $\pm \, \sigma$ \\
\hline
tuesday   &    0.000 $\pm$  0.000 \\
wednesday &     0.052 $\pm$  0.013 \\
thursday  &     0.067 $\pm$  0.023 \\
friday    &     0.437 $\pm$  0.127 \\
monday    &     0.538 $\pm$  0.057 \\
area      &     0.551 $\pm$  0.241 \\
january   &     0.569 $\pm$  0.132 \\
breaks    &     0.569 $\pm$  0.165 \\
february  &     0.573 $\pm$  0.112 \\
sites     &     0.576 $\pm$  0.195 \\
\hline
saturday  &     1.497 $\pm$  0.085 \\
sunday    &     1.526 $\pm$  0.070 \\
\hline
\end{tabular}
\caption*{Tuesday}
\label{Tab:Tuesday_UK}
}
\caption{Closest words (in the cosine sense) to "february","snow" and "tuesday" for the UK}
\label{Tab:cosine_words_UK1}
\end{table}

\begin{table}[H]
\parbox{.3\linewidth}{
\centering
\begin{tabular}{|c|c|}
\hline
Word &  dcos $\pm \, \sigma$ \\
\hline
février   &     0.000 $\pm$  0.000 \\
janvier   &     0.067 $\pm$  0.019 \\
décembre  &     0.124 $\pm$  0.030 \\
négatives &     0.181 $\pm$  0.070 \\
givrants  &     0.254 $\pm$  0.099 \\
glacial   &     0.274 $\pm$  0.110 \\
glaciales &     0.276 $\pm$  0.086 \\
gelées    &     0.312 $\pm$  0.088 \\
neige     &     0.338 $\pm$  0.147 \\
flocons   &     0.348 $\pm$  0.106 \\
\hline
juillet   &     1.626 $\pm$ 0.074 \\
août      &     1.459 $\pm$ 0.080 \\
\hline
\end{tabular}
\caption*{February}
\label{Tab:February_FR}
}
\hfill
\parbox{.3\linewidth}{
\centering
\begin{tabular}{|c|c|}
\hline
Word &  dcos $\pm \, \sigma$ \\
\hline
neige     &     0.000 $\pm$  0.000 \\
flocons   &     0.278 $\pm$  0.119 \\
négatives &     0.312 $\pm$  0.115 \\
givrants  &     0.324 $\pm$  0.139 \\
février   &     0.338 $\pm$  0.147 \\
glacial   &     0.338 $\pm$  0.094 \\
glaciales &     0.347 $\pm$  0.113 \\
froides   &     0.385 $\pm$  0.125 \\
gelées    &     0.407 $\pm$  0.089 \\
janvier   &     0.410 $\pm$  0.149 \\
\hline
août   &     1.262 $\pm$  0.146 \\
chaud   &     1.179 $\pm$  0.161 \\
\hline
\end{tabular}
\caption*{Snow}
\label{Tab:Snow_FR}
}
\hfill
\parbox{.3\linewidth}{
\centering
\begin{tabular}{|c|c|}
\hline
Word & dcos $\pm \, \sigma$ \\
\hline
mardi    &     0.000 $\pm$  0.000 \\
jeudi    &     0.103 $\pm$  0.058 \\
vendredi &     0.119 $\pm$  0.041 \\
mercredi &     0.126 $\pm$  0.024 \\
lundi    &     0.369 $\pm$  0.071 \\
fortes   &     0.594 $\pm$  0.191 \\
surface  &     0.644 $\pm$  0.167 \\
général  &     0.697 $\pm$  0.139 \\
saint    &     0.699 $\pm$  0.253 \\
vaste    &     0.714 $\pm$  0.106 \\
\hline
samedi   &     1.323 $\pm$  0.124 \\
dimanche   &    1.401 $\pm$  0.157 \\
\hline
\end{tabular}
\caption*{Tuesday}
\label{Tab:Tuesday_FR}
}
\caption{Closest words (in the cosine sense) to "february","snow" and "tuesday" for France}
\label{Tab:cosine_words_FR1}
\end{table}

\begin{figure}[H]
    \centering
    \begin{subfigure}[b]{0.49\textwidth}
        \includegraphics[width=\textwidth]{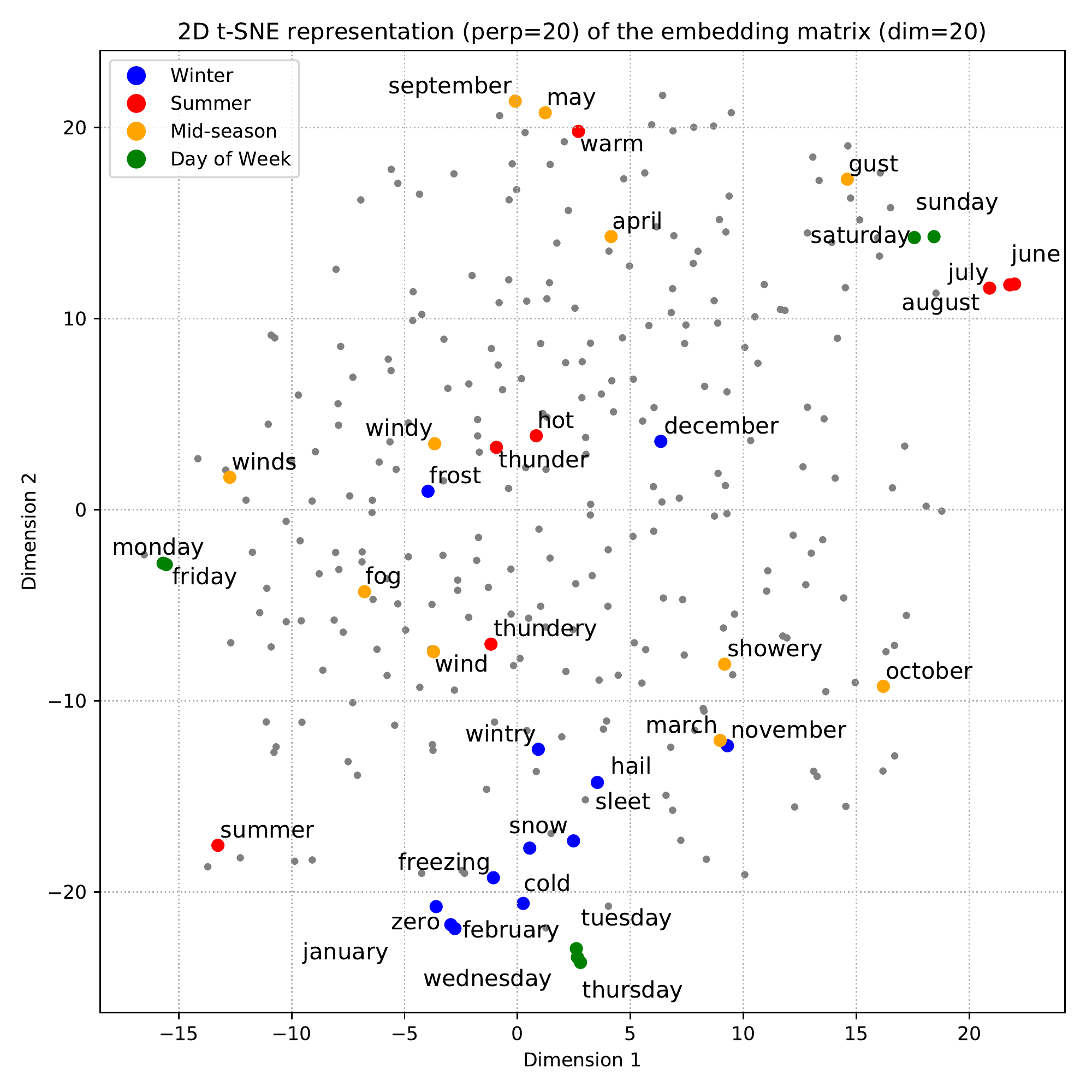}
        \caption{UK}
    \end{subfigure}
    \hspace{-0.2cm} 
    \begin{subfigure}[b]{0.49\textwidth}
        \includegraphics[width=\textwidth]{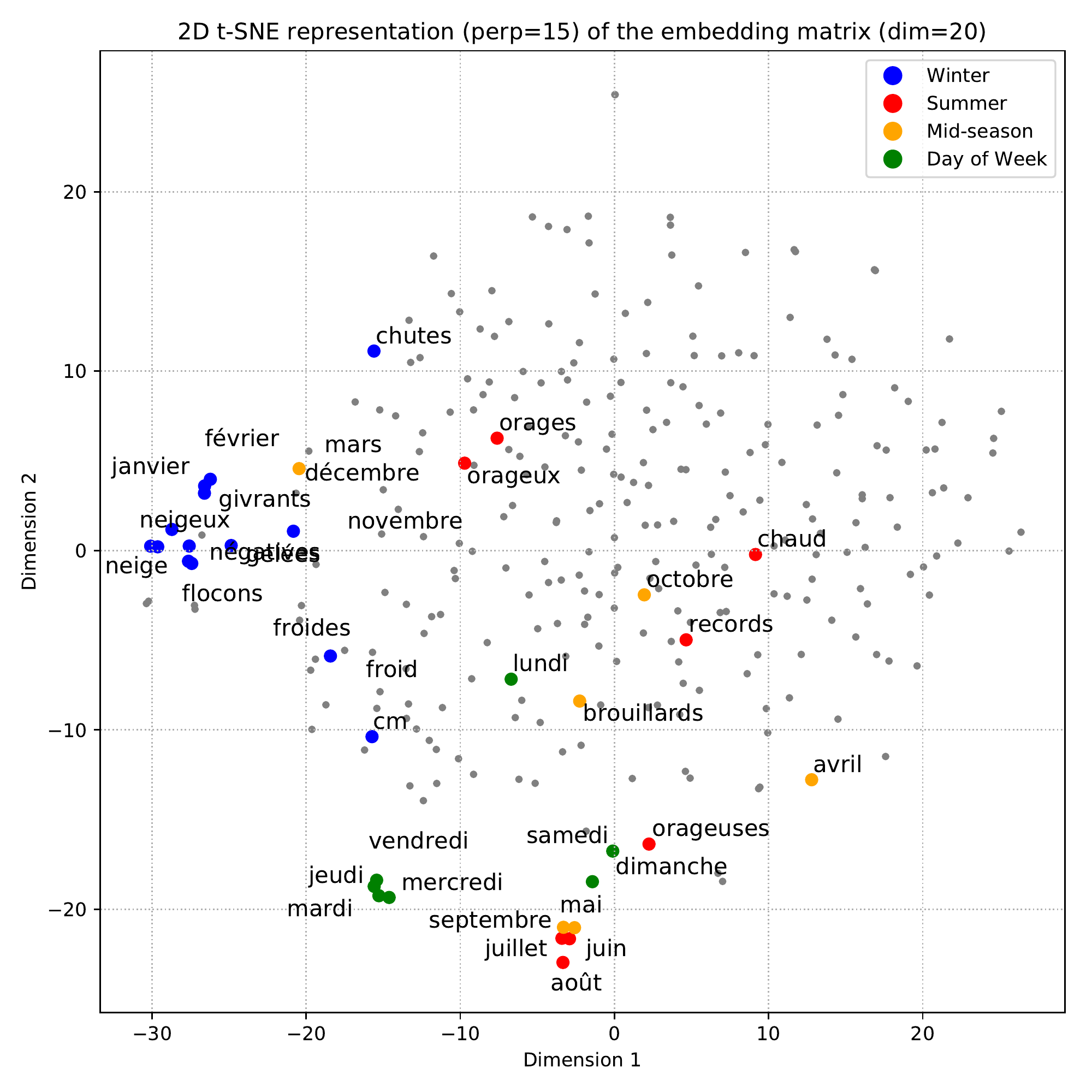}
        \caption{France}
    \end{subfigure}
    \caption{2D t-SNE projections of the embedding matrix for both languages.} 
    \label{Fig:tSNE}
\end{figure}

In order to achieve a global view of the embeddings, the t-SNE algorithm \cite{maaten2008visualizing} is applied to project an embedding matrix into a 2 dimensional space, for both languages. The observations for the few aforementioned words are confirmed by this representation, as plotted in figure \ref{Fig:tSNE}. Thematic clusters can be observed, roughly corresponding to winter, summer, week-days, week-end days for both languages. Globally summer and winter seem opposed, although one should keep in mind that the t-SNE representation does not preserve the cosine distance. The clusters of the French embedding appear much more compact than the UK one, comforting the observations made when explicitly calculating the cosine distances.
 
\section{Conclusion}
\label{Sec:Conclusion}

In this study, a novel pipeline to predict three types of time series using exclusively a textual source was proposed. Making use of publicly available daily weather reports, we were able to predict the electricity consumption with less than 5\% of MAPE for both France and the United-Kingdom. Moreover our average national temperature and wind speed predictions displayed sufficient accuracy to be used to replace missing data or as first approximation in traditional models in case of unavailability of meteorological features.

The texts were encoded numerically using either TF-IDF or our own neural word embedding. A plethora of machine learning algorithms such as random forests or neural networks were applied on top of those representations. Our results were consistent over language, numerical representation of the text and prediction algorithm, proving the intrinsic value of the textual sources for the three considered time series. Contrarily to previous works in the field of textual data for time series forecasting, we went in depth and quantified the impact of words on the variations of the series. As such we saw that all the algorithms naturally extract calendar and meteorological information from the texts, and that words impact the time series in the expected way (e.g. winter words increase the consumption and summer ones decrease it). Despite being trained on a regular quadratic loss, our neural word embedding spontaneously builds geometric properties. Not only does the norm of a word vector reflect its significance, but the words are also grouped by topic with for example winter, summer or day of the week clusters.

Note that this study was a preliminary work on the use of textual information for time series prediction, especially electricity demand one. The long-term goal is to include multiple sources of textual information to improve the accuracy of state-of-the-art methods or to build a text based forecaster which can be used to increase the diversity in a set of experts for electricity consumption \cite{gaillard2015forecasting}. However due to the redundancy of the information of the considered weather reports with meteorological features, it may be necessary to consider alternative textual sources. The use of social media such as Facebook, Twitter or Instagram may give interesting insight and will therefore be investigated in future work.

\appendix

\section{}

\begin{figure}[H]
    \centering
    \includegraphics[scale=0.35]{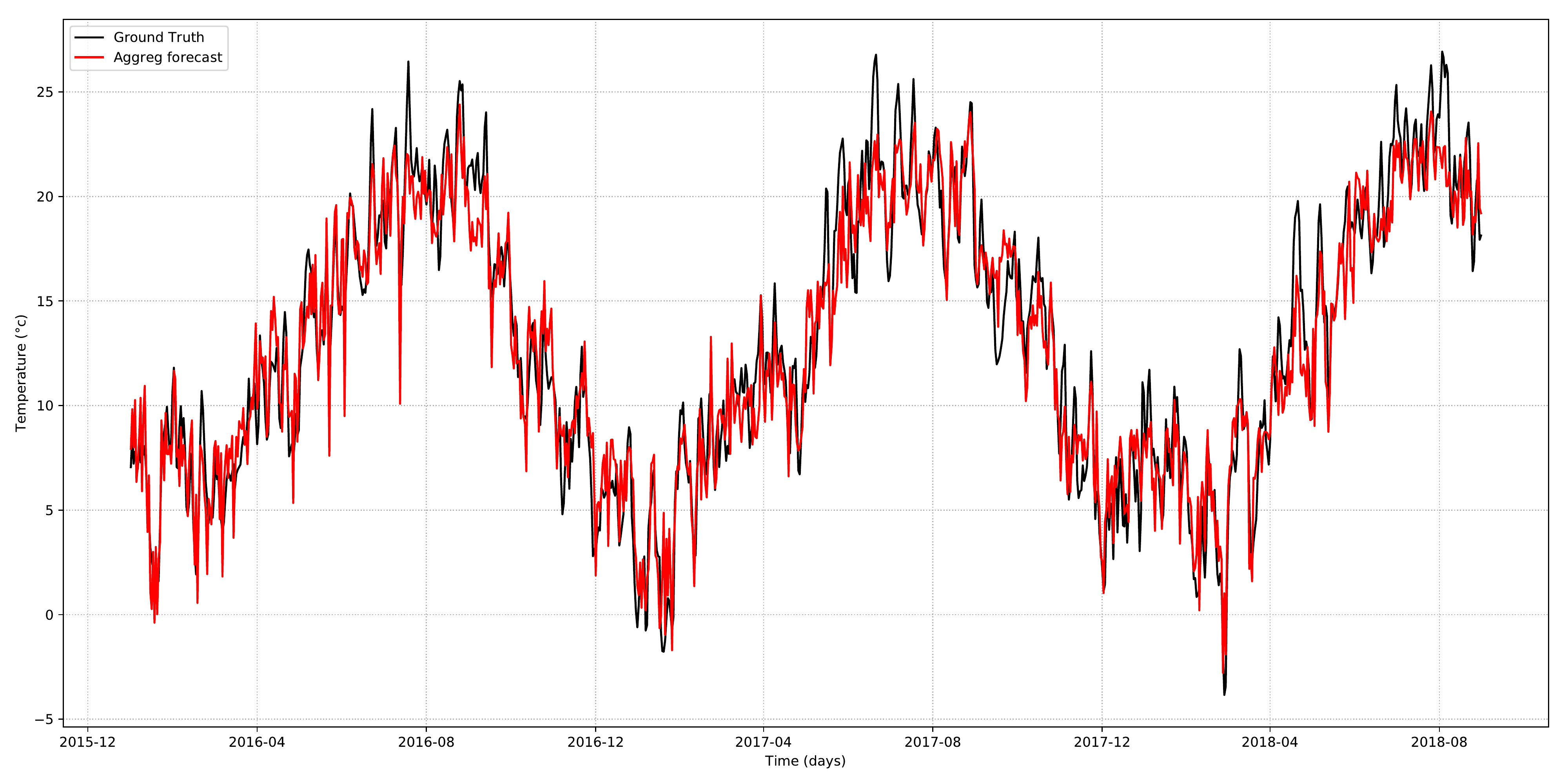}
    \caption{Overlapping of prediction and national Temperature (France)}
    \label{Fig:overlap_FR_Temp}
\end{figure}

Additional results for the prediction tasks on temperature and wind speed can be found in tables \ref{Tab:err_temp_FR} to \ref{Tab:err_vent_UK}. An example of forecast for the French temperature is given in figure \ref{Fig:overlap_FR_Temp}.

\begin{table}[H]
    \centering
    \begin{tabular}{|c|c|c|c|c|c|}
        \hline
        Method & Encoding & MAPE (\%) & RMSE (\degree c) & MAE (\degree c) & $R^2$ \\
        \hline
        RF & \multirow{4}{*}{TF-IDF} & $ 15.67 \pm 0.05$ & $ 2.32 \pm 0.01 $ & $1.78 \pm 0.01$ & $0.88 \pm 0.00$ \\
        RF (Sel) & & $15.28 \pm 0.08$ & $2.23 \pm 0.01$ & $1.73 \pm 0.01$ & $0.89 \pm 0.00 $ \\
        MLP &  & $14.50 \pm 0.28$ & $1.96 \pm 0.03$ & $ 1.53 \pm 0.03 $ & $0.91 \pm 0.00$ \\
        LASSO &  & $20.48 \pm 0.00$ & $2.57 \pm 0.00$ & $2.05 \pm 0.00$ & $0.85\pm 0.00$ \\
        \hline
        CNN & \multirow{2}{*}{Embedding} & $16.55 \pm 0.17 $ & $2.30 \pm 0.03 $ & $ 1.80 \pm 0.02$ & $ 0.88 \pm 0.01$ \\
        RNN & & $14.60 \pm 0.35$ & $1.95 \pm 0.03$ & $1.53\pm 0.03 $ & $0.91 \pm 0.00$ \\
        \hline
        Agg (RNN+MLP) & Both & $ 13.89 \pm 0.18$ & $1.86 \pm 0.01$ & $1.47 \pm 0.01$ & $ 0.92 \pm 0.00 $ \\ 
        \hline
    \end{tabular}
    \caption{Forecast errors on the national temperature for France.}
    \label{Tab:err_temp_FR}
\end{table}


\begin{table}[H]
    \centering
    \begin{tabular}{|c|c|c|c|c|c|}
        \hline
        Method & Encoding & MAPE (\%) & RMSE ($\text{m}.\text{s}^{-1}$) & MAE ($\text{m}.\text{s}^{-1}$) & $R^2$ \\
        \hline
        RF & \multirow{4}{*}{TF-IDF} & $ 18.68 \pm 0.05$ & $ 0.72 \pm 0.00 $ & $0.56 \pm 0.0$ & $0.44 \pm 0.00$ \\
        RF (Sel) & & $19.25 \pm 0.05$ & $0.71 \pm 0.00$ & $0.57 \pm 0.00$ & $0.45 \pm 0.00 $ \\
        MLP &  & $17.65 \pm 0.28$ & $0.70 \pm 0.01$ & $ 0.54 \pm 0.01 $ & $0.47 \pm 0.02$ \\
        LASSO &  & $18.61 \pm 0.00$ & $0.71 \pm 0.00$ & $0.55 \pm 0.00$ & $0.46\pm 0.00$ \\
        \hline
        CNN & \multirow{2}{*}{Embedding} & $20.36 \pm 1.49 $ & $0.72 \pm 0.04 $ & $ 0.59 \pm 0.03$ & $ 0.43 \pm 0.06$ \\
        RNN & & $18.11 \pm 0.63$ & $0.68 \pm 0.01$ & $0.53\pm 0.01 $ & $0.51 \pm 0.02$ \\
        \hline
        Agg (RNN+MLP) & Both & $ 17.29 \pm 0.35$ & $0.66 \pm 0.01$ & $0.52 \pm 0.01$ & $ 0.53 \pm 0.01 $ \\ 
        \hline
    \end{tabular}
    \caption{Forecast errors on the national wind for France.}
    \label{Tab:err_vent_FR}
\end{table}

\begin{table}[H]
    \centering
    \begin{tabular}{|c|c|c|c|c|c|}
        \hline
        Method & Encoding & MAPE (\%) & RMSE (\degree c) & MAE (\degree c) & $R^2$ \\
        \hline
        RF & \multirow{4}{*}{TF-IDF} & $ 16.51\pm 0.06$ & $1.67\pm 0.00$ & $1.33\pm 0.00$ & $ 0.88\pm 0.00$ \\
        RF (Sel) & & $16.87 \pm 0.05$ & $1.69\pm 0.00$ & $ 1.35\pm 0.00$ & $0.87\pm 0.00$ \\
        MLP &  & $18.11\pm 0.35$ & $1.81\pm 0.05$ & $1.45\pm 0.04$ & $0.85\pm 0.01$ \\
        LASSO &  & $19.20 \pm 0.00$ & $1.87 \pm 0.00$ & $1.49\pm 0.00$ & $0.84\pm 0.00$ \\
        \hline
        CNN & \multirow{2}{*}{Embedding} & $17.48 \pm 0.57$ & $1.80\pm 0.09$ & $1.41\pm 0.08$ & $ 0.86\pm 0.02$ \\
        RNN & & $15.80\pm 0.68$ & $1.58\pm 0.05$ & $1.24\pm 0.04$ & $0.89\pm 0.01$ \\
        \hline
        Agg (RNN+RF) & Both & $15.06\pm 0.18$ & $1.53\pm 0.02$ & $1.21 \pm 0.02$ & $ 0.89\pm 0.00$ \\ 
        \hline
    \end{tabular}
    \caption{Forecast errors on the national temperature for Great-Britain.}
    \label{Tab:err_temp_UK}
\end{table}

\begin{table}[H]
    \centering
    \begin{tabular}{|c|c|c|c|c|c|}
        \hline
        Method & Encoding & MAPE (\%) & RMSE ($\text{m}.\text{s}^{-1}$) & MAE ($\text{m}.\text{s}^{-1}$) & $R^2$ \\
        \hline
        RF & \multirow{4}{*}{TF-IDF} & $ 19.80\pm 0.03$ & $1.00\pm 0.00$ & $0.78\pm 0.00$ & $ 0.52\pm 0.00$ \\
        RF (Sel) & & $20.08 \pm 0.04$ & $1.00\pm 0.00$ & $ 0.79\pm 0.00$ & $0.52\pm 0.00$ \\
        MLP &  & $21.01\pm 0.57$ & $1.03\pm 0.01$ & $0.82\pm 0.01$ & $0.49\pm 0.01$ \\
        LASSO &  & $19.76 \pm 0.00$ & $0.99 \pm 0.00$ & $0.78\pm 0.00$ & $0.53\pm 0.00$ \\
        \hline
        CNN & \multirow{2}{*}{Embedding} & $20.61 \pm 0.42$ & $1.02\pm 0.01$ & $0.80\pm 0.01$ & $ 0.50\pm 0.01$ \\
        RNN & & $20.58\pm 0.41$ & $1.07\pm 0.02$ & $0.83\pm 0.01$ & $0.45\pm 0.02$ \\
        \hline
        Agg (LASSO+RF) & TF-IDF & $19.31\pm 0.02$ & $0.96\pm 0.00$ & $0.76 \pm 0.00$ & $ 0.56\pm 0.00$ \\ 
        \hline
    \end{tabular}
    \caption{Forecast errors on the national wind for Great-Britain.}
    \label{Tab:err_vent_UK}
\end{table}

While not strictly normally distributed, the residuals for the French electricity demand display an acceptable behavior. This holds also true for the British consumption, and both temperature time series, but is of lesser quality for the wind one.

\begin{figure}[H]
    \centering
    \includegraphics[scale=0.54]{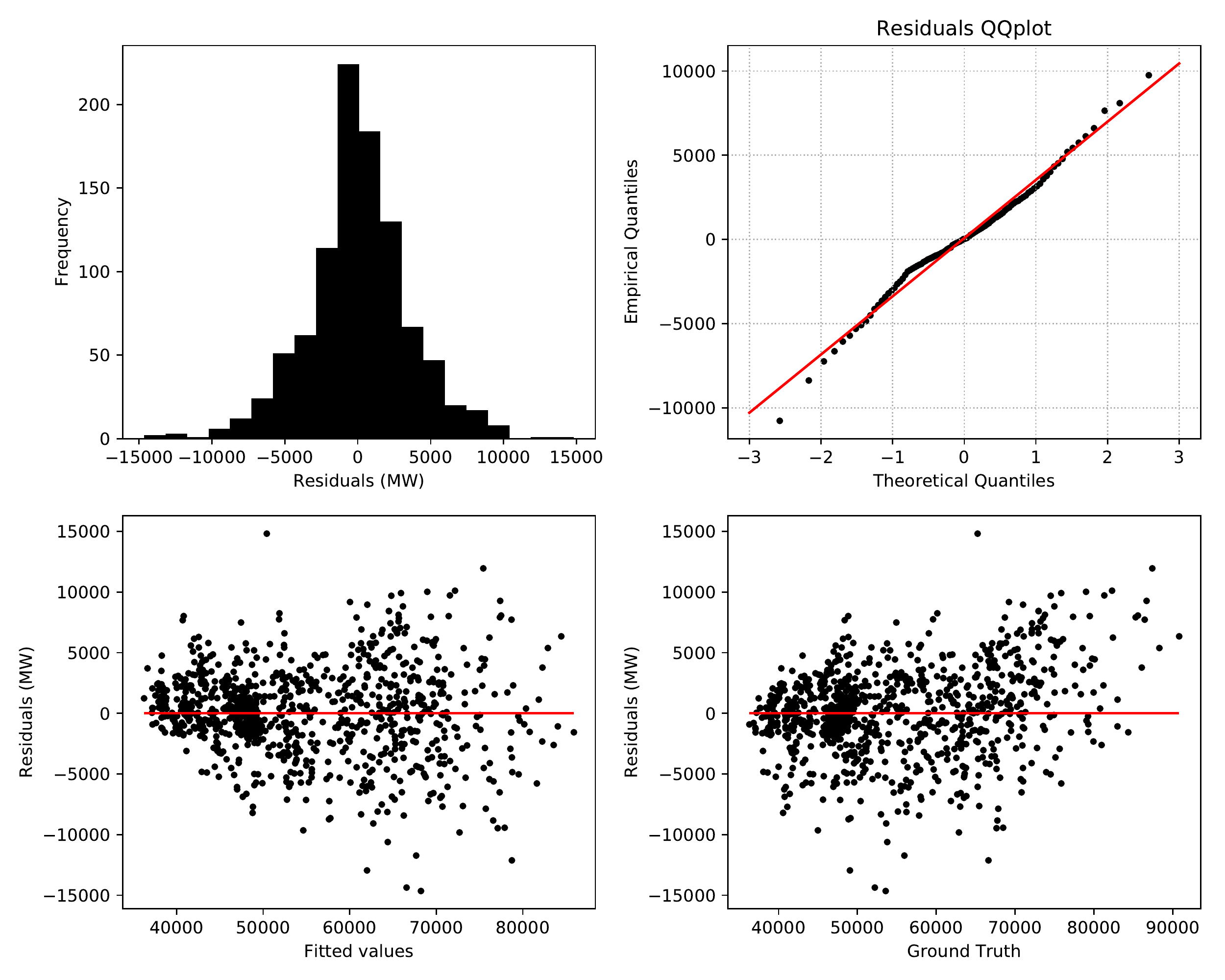}
    \caption{Residual analysis of the French aggregated predictor.}
    \label{Fig:FR_residuals_analysis}
\end{figure}

\begin{figure}[H]
    \centering
    \includegraphics[scale=0.45]{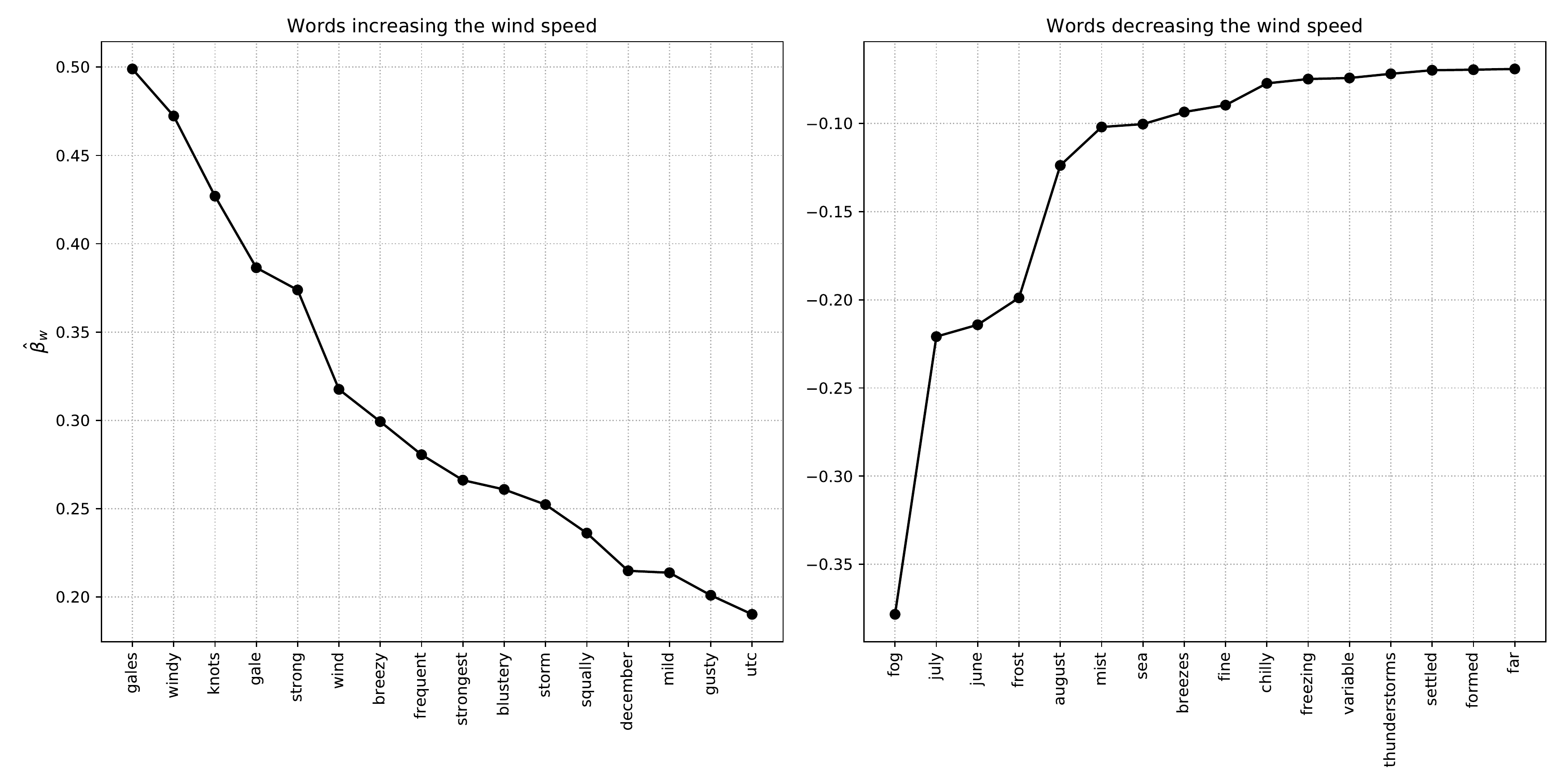}
    \caption{Coefficients $\beta_w$ in the British wind LASSO regression.}
    \label{Fig:LASSO_coeffs_wind_UK}
\end{figure}

The the UK wind LASSO regression, the words with the highest coefficients $\beta_w$ are indeed related to strong wind phenomena, whereas antagonistic ones such as "fog" or "mist" have strongly negative ones as expected (fig. \ref{Fig:LASSO_coeffs_wind_UK}).

\begin{figure}[H]
    \centering
    \begin{subfigure}[b]{0.49\textwidth}
        \includegraphics[width=\textwidth]{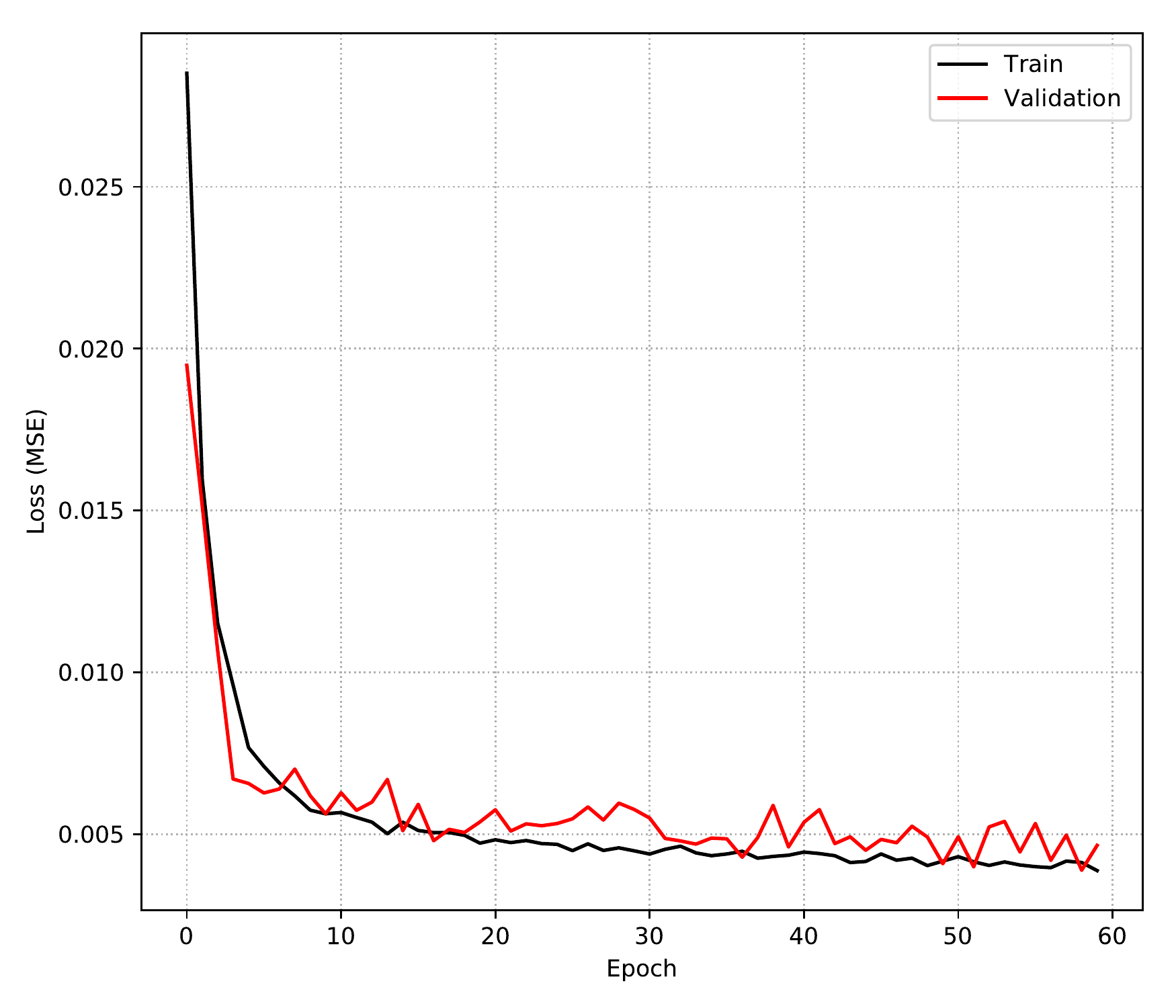}
        \caption{France}
    \end{subfigure}
    \hspace{-0.2cm} 
    \begin{subfigure}[b]{0.49\textwidth}
        \includegraphics[width=\textwidth]{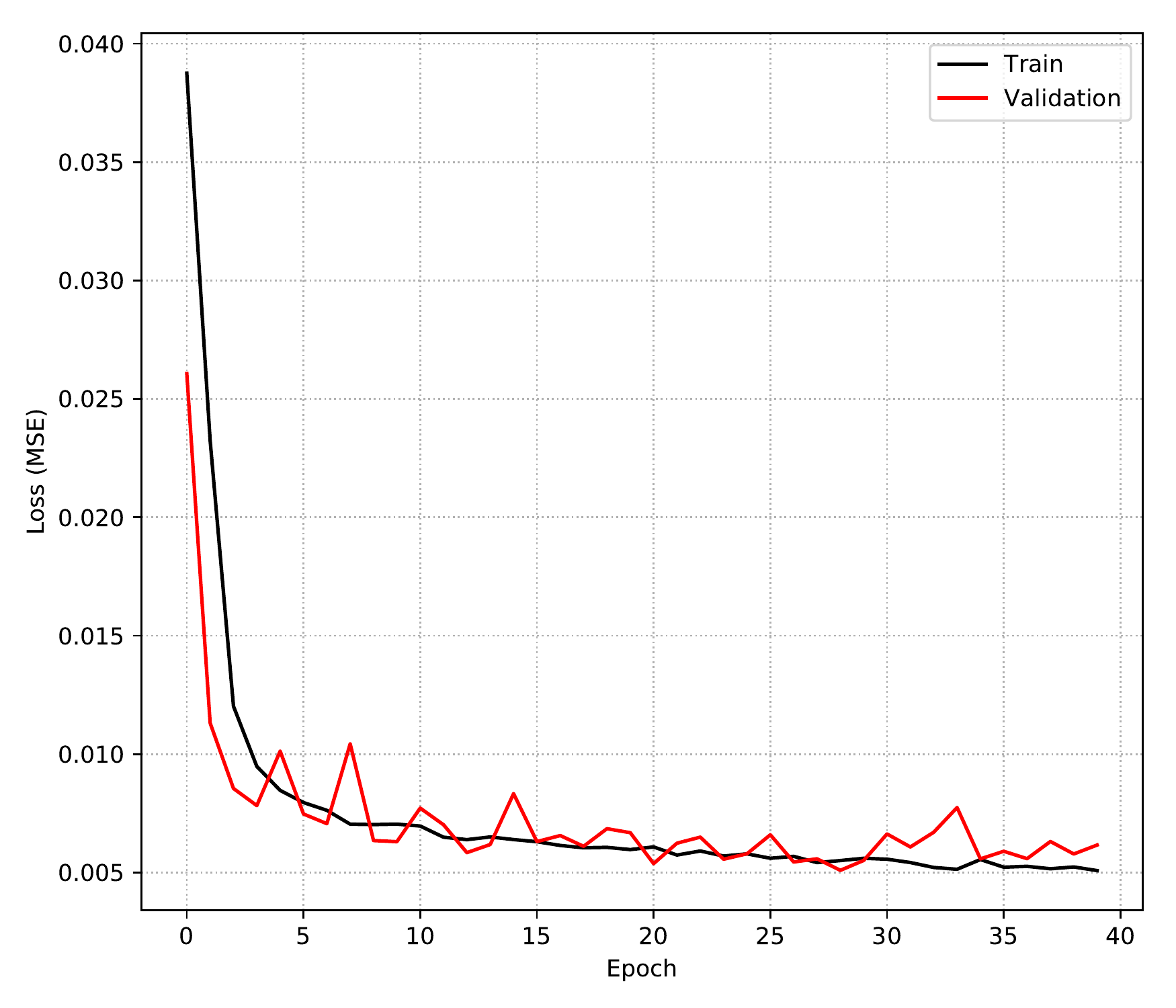}
        \caption{UK}
    \end{subfigure}
    \caption{Loss (Mean Squared Error) evolution of the electricity demand RNN for both languages.} \label{Fig:Loss_RNN}
\end{figure}

For both languages we represented the evolution of the (normalized) losses for the problem of load regression in fig. \ref{Fig:Loss_RNN}. The aspect is a typical one, with the validation loss slightly above the training one. The slightly erratic behavior of the former one is possibly due to a lack of data to train the embeddings.

The cosine distances for three other major words and for both corpora have been calculated as well. The results are given in tables \ref{Tab:cosine_words_UK2} and \ref{Tab:cosine_words_FR2}. For both languages, the three summer months are grouped together, and so are the two week-end days. However again the results are less clear for the English language. They are especially mediocre for "hot", considering that only "warm" seems truly relevant and that "August" for instance is quite far away. For the French language instead of "hot" the distances to "thunderstorms" were calculated. The results are quite satisfactory, with "orageux"/"orageuse" ("thundery") coming in the two first places and related meteorological phenomena ("cumulus" and "grêle", meaning "hail") relatively close as well. For the French case, Saturday and Sunday are very close to summer related words. This observation probably highlights the fact that the RNN groups load increasing and decreasing words in opposite parts of the embedding space.

\begin{table}[H]
\parbox{.3\linewidth}{
\centering
\begin{tabular}{|c|c|}
\hline
Word & dcos $\pm \, \sigma$ \\
\hline
august     &     0.000 $\pm$  0.000 \\
july       &     0.104 $\pm$  0.031 \\
june       &     0.175 $\pm$  0.053 \\
hazy       &     0.244 $\pm$  0.091 \\
days       &     0.388 $\pm$  0.078 \\
large      &     0.419 $\pm$  0.079 \\
new        &     0.450 $\pm$  0.186 \\
felt       &     0.484 $\pm$  0.138 \\
heaviest   &     0.495 $\pm$  0.142 \\
especially &     0.506 $\pm$  0.148 \\
\hline
cold       &     1.265 $\pm$ 0.146 \\
summer     &     0.689 $\pm$ 0.194 \\
\hline
\end{tabular}
\caption*{August}
\label{Tab:August_UK}
}
\hfill
\parbox{.3\linewidth}{
\centering
\begin{tabular}{|c|c|}
\hline
Word & dcos $\pm \, \sigma$ \\
\hline
sunday   &     0.000 $\pm$  0.000 \\
saturday &     0.235 $\pm$  0.029 \\
new      &     0.283 $\pm$  0.141 \\
hazy     &     0.336 $\pm$  0.191 \\
teens    &     0.394 $\pm$  0.202 \\
heaviest &     0.394 $\pm$  0.138 \\
summer   &     0.443 $\pm$  0.155 \\
may      &     0.443 $\pm$  0.118 \\
warm     &     0.470 $\pm$  0.128 \\
end      &     0.527 $\pm$  0.225 \\
\hline
monday   &     1.127 $\pm$ 0.147 \\
friday   &     1.289 $\pm$ 0.120 \\
\hline
\end{tabular}
\caption*{Sunday}
\label{Tab:Sunday_UK}
}
\hfill
\parbox{.3\linewidth}{
\centering
\begin{tabular}{|c|c|}
\hline
mot &  dcos $\pm \, \sigma$ \\
\hline
hot          &     0.000 $\pm$  0.000 \\
occasionally &     0.648 $\pm$  0.142 \\
warm         &     0.694 $\pm$  0.136 \\
wake         &     0.695 $\pm$  0.236 \\
sea          &     0.714 $\pm$  0.204 \\
particularly &     0.722 $\pm$  0.269 \\
rural        &     0.736 $\pm$  0.146 \\
bursts       &     0.750 $\pm$  0.199 \\
colder       &     0.752 $\pm$  0.126 \\
heaviest     &     0.777 $\pm$  0.201 \\
\hline
august       &     0.950 $\pm$ 0.168 \\
december     &     1.057 $\pm$ 0.267 \\
\hline
\end{tabular}
\caption*{Hot}
\label{Tab:Hot_UK}
}
\caption{Closest words (in the cosine sense) to "August","Sunday" and "Hot" for the UK}
\label{Tab:cosine_words_UK2}
\end{table}

\begin{table}[H]
\parbox{.3\linewidth}{
\centering
\begin{tabular}{|c|c|}
\hline
Word &  dcos $\pm \, \sigma$ \\
\hline
août      &     0.000 $\pm$  0.000 \\
juin      &     0.142 $\pm$  0.046 \\
juillet   &     0.152 $\pm$  0.053 \\
mai       &     0.177 $\pm$  0.060 \\
septembre &     0.212 $\pm$  0.071 \\
dimanche  &     0.493 $\pm$  0.300 \\
orageux   &     0.499 $\pm$  0.115 \\
heure     &     0.584 $\pm$  0.165 \\
vosges    &     0.621 $\pm$  0.149 \\
orageuse  &     0.633 $\pm$  0.161 \\
\hline
froid     &     1.182 $\pm$ 0.116 \\
chaud     &     1.066 $\pm$ 0.075 \\
\hline
\end{tabular}
\caption*{August}
\label{Tab:August_FR}
}
\hfill
\parbox{.3\linewidth}{
\centering
\begin{tabular}{|c|c|}
\hline
Word &  dcos $\pm \, \sigma$ \\
\hline
dimanche  &     0.000 $\pm$  0.000 \\
samedi    &     0.323 $\pm$  0.084 \\
mai       &     0.472 $\pm$  0.277 \\
août      &     0.493 $\pm$  0.300 \\
juin      &     0.541 $\pm$  0.303 \\
septembre &     0.549 $\pm$  0.273 \\
juillet   &     0.571 $\pm$  0.281 \\
dessus    &     0.575 $\pm$  0.206 \\
instables &     0.586 $\pm$  0.105 \\
advection &     0.618 $\pm$  0.268 \\
\hline
lundi     &     1.094 $\pm$  0.208 \\
vendredi  &     1.370 $\pm$  0.164 \\
\hline
\end{tabular}
\caption*{Sunday}
\label{Tab:Sunday_FR}
}
\hfill
\parbox{.3\linewidth}{
\centering
\begin{tabular}{|c|c|}
\hline
Word &  dcos $\pm \, \sigma$ \\
\hline
orages      &     0.000 $\pm$ 0.000 \\
orageuse    &     0.540 $\pm$  0.193 \\
orageux     &     0.547 $\pm$  0.221 \\
heure       &     0.603 $\pm$  0.204 \\
cumulus     &     0.662 $\pm$  0.258 \\
grêle       &     0.662 $\pm$  0.208 \\
dessus      &     0.689 $\pm$  0.195 \\
septembre   &     0.701 $\pm$  0.196 \\
juillet     &     0.703 $\pm$  0.170 \\
supérieures &     0.704 $\pm$  0.194 \\
\hline
août        &     0.811 $\pm$ 0.119 \\
janvier     &     1.187 $\pm$ 0.084 \\
\hline
\end{tabular}
\caption*{Thunderstorms}
\label{Tab:Thunderstorms_FR}
}
\caption{Closest words (in the cosine sense) to "August","Sunday and "thunderstorms" for the France}
\label{Tab:cosine_words_FR2}
\end{table}


\newpage


\bibliographystyle{elsarticle-num} 
\bibliography{biblio.bib}



\end{document}